\documentclass[10pt,twocolumn,letterpaper]{article}

\usepackage{iccv}      

\usepackage{multirow}

\newcommand{\dcheck}[1]{{\color{black}#1}}

\newcommand{\lens}{\texttt{Lens}}

\newcommand{\adsqa}{\texttt{AdsQA}}

\newcommand{\mhref}[3][black]{\href{#2}{\color{#1}{#3}}}

\definecolor{iccvblue}{rgb}{0.21,0.49,0.74}
\usepackage[pagebackref,breaklinks,colorlinks,allcolors=iccvblue]{hyperref}
\usepackage{multirow}
\usepackage{makecell}
\usepackage{colortbl}
\usepackage{xcolor}
\usepackage{bbding}
\usepackage{graphicx}

\usepackage[most]{tcolorbox}
\tcbuselibrary{breakable}
\usepackage[x11names, svgnames, table]{xcolor}

\def\paperID{*****} 
\def\confName{ICCV}
\def\confYear{2025}

\title{MARS2 2025 Challenge on Multimodal Reasoning: \\ Datasets, Methods, Results, Discussion, and Outlook}

\author{Peng Xu\textsuperscript{\dag} 
\and
Shengwu Xiong\textsuperscript{\dag}
\and
Jiajun Zhang\textsuperscript{\dag}
\and
\vspace{0.8pt}
Yaxiong Chen\textsuperscript{\dag}
\and
Bowen Zhou\textsuperscript{\ddag}
~~
Chen Change Loy\textsuperscript{\ddag}
~~
David A. Clifton\textsuperscript{\ddag}
~~
Kyoung Mu Lee\textsuperscript{\ddag}
~~
Luc Van Gool\textsuperscript{\ddag}
\and
Ruiming He\textsuperscript{a,b,c}
% ...
\\
{\small \textsuperscript{\dag} competition organizers; 
\textsuperscript{\ddag} steering committee;}\\
{\small \textsuperscript{a} organizing contributors; \textsuperscript{b} dataset contributors; \textsuperscript{c} baseline implementors; \textsuperscript{d} competition participants;} \\
{\small The authors’ teams and affiliations are in the appendix.} \\
{\small ICCV 2025 MARS2 workshop \& challenge website: 
\url{https://mars2workshop.github.io/iccv2025/} } \\
{\small MARS2 GitHub organization: \url{https://github.com/mars2workshop}
}\\
}

\begin{document}
\maketitle

\begin{abstract}
This paper reviews the MARS2 2025 Challenge on Multimodal Reasoning.
We aim to bring together different approaches in multimodal machine learning and LLMs via a large benchmark. We hope it better allows researchers to follow the state-of-the-art in this very dynamic area.
Meanwhile, a growing number of testbeds have boosted the evolution of general-purpose large language models. 
Thus, this year’s MARS2 focuses on real-world and specialized scenarios to broaden the multimodal reasoning applications of MLLMs.
Our organizing team released two tailored datasets \lens{} and \adsqa{} as test sets, which support \dcheck{general reasoning} in 12 daily scenarios and \dcheck{domain-specific reasoning} in advertisement videos, respectively.  
We evaluated \dcheck{40+} baselines that include both generalist MLLMs and task-specific models, and opened up three competition tracks, \ie, Visual Grounding in Real-world Scenarios (VG-RS), Visual Question Answering with Spatial Awareness (VQA-SA), and Visual Reasoning in Creative Advertisement Videos (VR-Ads). 
Finally, \dcheck{76} teams from the renowned academic and industrial institutions have registered and \dcheck{40+} valid submissions (out of \dcheck{1200+}) have been included in our ranking lists.
Our datasets, code sets (\dcheck{40+} baselines and \dcheck{15+} participants' methods), and rankings are publicly available on the MARS2 workshop website and our GitHub organization page \url{https://github.com/mars2workshop/}, where our updates and announcements of upcoming events will be continuously provided. 
\end{abstract}

\section{Introduction}
\label{sec:intro}

Large language models (LLMs)~\cite{naveed2023comprehensive,minaee2024large} represent a major advance and may mark a crucial step towards Artificial General Intelligence (AGI). 
LLMs exhibit various characteristics that push the envelope of traditional deep learning, including generalization and emergent abilities~\cite{berti2025emergent}, chain-of-thought (CoT)~\cite{wei2022chain}, instruction understanding and following~\cite{zhou2023instruction}, tool calling~\cite{schick2023toolformer}, \etc.
Specifically, LLMs equipped with long chain reasoning capabilities such as OpenAI o1~\cite{jaech2024openai} and DeepSeek-R1~\cite{guo2025deepseekr1} have opened up a new avenue of large reasoning models.
Furthermore, multimodal large language models (MLLMs) have expanded the application boundaries of language models and further advanced multimodal machine learning~\cite{xu2023multimodal,xu2025guest}.

Given the remarkable progress of LLMs and multimodal machine learning, a natural question for researchers is what to focus on next. Simply deploying LLMs on tasks that conventional deep learning already solves well is unsurprising and adds little novelty. 
Instead, we should pursue more challenging~\cite{besta2024graph} and specialized \cite{shen2024tag} tasks  
that require complex multimodal reasoning and slower, ``System 2'' thinking~\cite{evans2003two}, borrowing from Kahneman's dual-process view of mind~\cite{kahneman2011thinking}).
Thus, we have organized the MARS2 Workshop and Challenge ``Multimodal Reasoning and Slow Thinking in the Large Model Era: Towards System 2 and Beyond''.
The goal is to bring together perspectives from multiple disciplines and to help researchers follow the state-of-the-art in both multimodal machine learning and multimodal large language models.

\paragraph{Motivation and objectives.}

Domain-specific problems highlight the challenge of equipping general-purpose models with specialized expertise.
Accordingly, this year’s MARS2 focuses on real-world and specialized scenarios to broaden the multimodal reasoning applications of MLLMs.
In particular, we explore the following challenges:
\begin{itemize}
    \item \textbf{Synergistic effects among reasoning tasks.} 
    \dcheck{Existing reasoning benchmarks fail to guarantee that different task samples come from the same data distribution, thus falling short of evaluating the synergistic effects among the different reasoning tasks \cite{yao2025lens}.}
    \item \textbf{Non-stepwise complex reasoning.}  \dcheck{It has been observed~\cite{zhang2024accessing,guo2024deepseekcoder} that LLMs already work well for step-wise problems such as math and programming that are based on explicit theorems and programming syntax, compatible with the chain-style reasoning approach \texttt{``if A, then B''}.
    To further advance LLM reasoning, we should study complex reasoning that goes beyond step-wise approaches.}
\end{itemize}

Therefore, we released two tailored datasets \lens~\cite{yao2025lens} and \adsqa~\cite{adsqa_2025_ICCV} as our competition test sets to draw the attention of the community to these two issues.
For details, we refer to Section~\ref{sec:datasets}.

\paragraph{Overview and features.}

This challenge was organized in conjunction with the ICCV 2025 MARS2 Workshop ``Multimodal Reasoning and Slow Thinking in the Large Model Era: Towards System 2 and Beyond''.

Our committee organized a team consisting of \dcheck{$80+$} data contributors and baseline model testers.
Two large-scale and tailored datasets \texttt{Lens}~\cite{yao2025lens} and \texttt{AdsQA}~\cite{adsqa_2025_ICCV} were released as test sets, which support \dcheck{general reasoning} in 12 daily scenarios and \dcheck{domain-specific reasoning} in advertisement videos, respectively.  
We evaluated over 40 baselines that include both generalist MLLMs and task-specific models, covering open-source and commercial models.
This year, we provided three competition tracks: Track \#1 Visual Grounding in Real-world Scenarios (VG-RS), Track \#2 Visual Question Answering with Spatial Awareness (VQA-SA), and Track \#3 Visual Reasoning in Creative Advertisement Videos (VR-Ads). 
The challenge was divided into two stages and lasted for more than two months. 
\dcheck{76} teams from renowned academic and industrial institutions (\eg, \dcheck{ByteDance, Meituan, NVIDIA, and Samsung}) have registered and submitted \dcheck{1200+} entries in total, of which \dcheck{40+} valid submissions have been included in our final ranking lists.
Our datasets, codes (\dcheck{40+} baselines and \dcheck{15+} participants' methods), and rankings are publicly available at the MARS2 workshop website and our GitHub repositories.

Thanks to the significant efforts of all contributors and participants, the MARS2 workshop met our expectations. This year's challenge boasts:
\begin{itemize}
    \item {\bf Tailored datasets.} Our datasets \texttt{Lens} and \texttt{AdsQA} come with distinct characteristics. For example, 
    the samples of \texttt{Lens} were manually collected from social media, with $53\%$ posted after January 2025.
    Each sample supports Track \#1 and Track \#2, enabling the study of synergistic effects of multimodal reasoning tasks. 
    On the other hand, \texttt{AdsQA}
    is a challenging advertisement Video QA test set, and to the best of our knowledge, is the first attempt to use ad videos for tasks that evaluate LLMs. For more details of \texttt{Lens} and \texttt{AdsQA}, the reader is referred to Section~\ref{sec:datasets}. 
    
    \item {\bf Comprehensive benchmarks.} Our contributors evaluated \dcheck{40+} baselines involving both generalist MLLMs and task-specific models.
    The participants contributed \dcheck{$40+$} valid solutions, by using not only open-source models like Qwen~\cite{bai2023qwen} but also some commercial LLMs like \dcheck{TeleMM}\footnote{\url{https://www.teleai.com/}}.
    The sizes of the baselines and the participants' models range from \dcheck{3B to 72B}. 
    \dcheck{This benchmark provides a comprehensive comparison covering ``generalist models \vs specialist models'', ``large models \vs small models'', ``open-source models \vs closed-source models'', \etc.}
    
    \item {\bf Open-source reproducibility.} Our committee has set up a GitHub organization 
\url{https://github.com/mars2workshop} and a dedicated webpage \url{https://mars2workshop.github.io/iccv2025/}, where our datasets and codes (\dcheck{40+} baselines and \dcheck{15+} participants' methods) are  publicly available. This should ensure reproducibility.

    \item {\bf Post-competition discussion.} After the submission deadline, the MARS2 committee has initiated multiple rounds of discussion, inviting additional contributors and covering interesting issues like hard samples, corner cases, and valid tricks. In particular, we have reopened submissions to allow participants to conduct  ablation studies. 
\end{itemize}

\paragraph{Organization of this report.}

The rest of this report is organized as follows. 
Section \ref{sec:challenge} provides the overview of the MARS2 2025 challenge, introducing tailored datasets (Section~\ref{sec:datasets}), competition tracks (Section~\ref{sec:tacks}), evaluation protocols (Section~\ref{sec:protocols}), and other details (Section~\ref{sec:rules_phases}).
Section \ref{sec:methods_results} presents the solutions and baselines, including methods, details, and rankings.
In Section \ref{sec:discussion_outlook}, we discuss open problems and potential research directions before 
this report concludes in Section \ref{sec:conclusion}.

\section{Challenge}
\label{sec:challenge}

\subsection{Tailored Datasets}
\label{sec:datasets}

Based on the motivation introduced in Section~\ref{sec:intro}, our committee collected two large-scale multimodal datasets \lens~\cite{yao2025lens} and \adsqa~\cite{adsqa_2025_ICCV} from real-world scenarios to explore the synergistic effects among the reasoning tasks and non-stepwise complex reasoning, respectively.

\subsubsection{\lens{} Dataset}

\paragraph{Statistics.}

\lens{} is created to support the multi-\underline{l}evel \underline{e}valuatio\underline{n} of multimodal rea\underline{s}oning.
It is a multi-level benchmark that provides three progressive task tiers, \ie, perception, understanding, and reasoning.
It has 3.4K images and 60K+ human-authored questions covering eight tasks (\eg, described object counting, region-wise OCR, and scene knowledge
inference) and 12 daily scenarios (\eg, streets, stations, schools, and homes). 

\paragraph{Features.}

\begin{figure*}[h] %
   \centering
   \begin{minipage}[b]{.47\linewidth}
      \includegraphics[width=1.\linewidth]{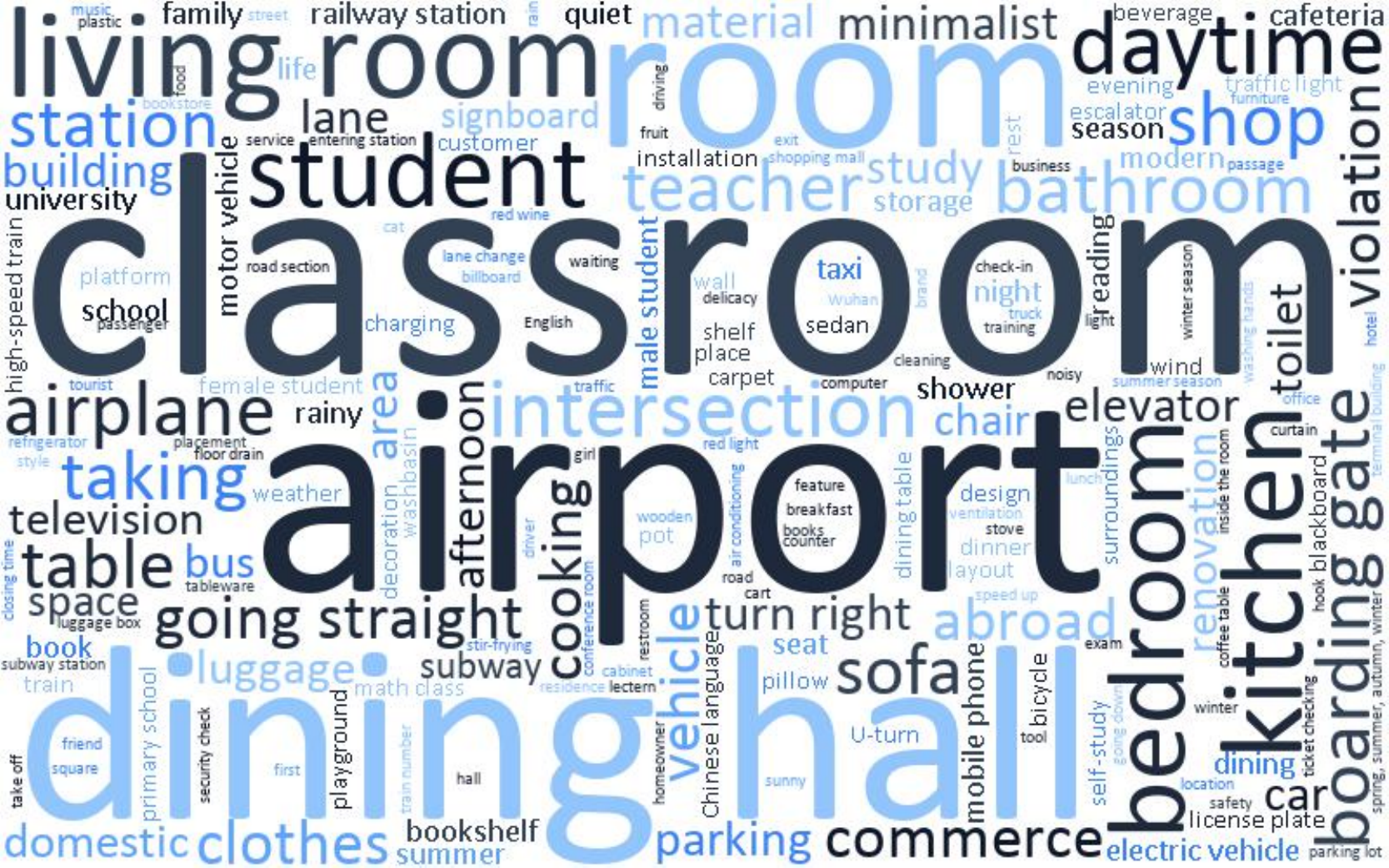}
      \caption{
      A word cloud illustrating the textual content of the \lens{} \cite{yao2025lens} dataset. The frequent scene-specific terms suggest a diversity of daily scenarios, while directional terms (\eg, \textit{turn right}) highlight its capacity for evaluating spatial reasoning.
      } \label{fig:lens-word-cloud}
   \end{minipage}\qquad
   \begin{minipage}[b]{.47\linewidth}
      \includegraphics[width=1.\linewidth]{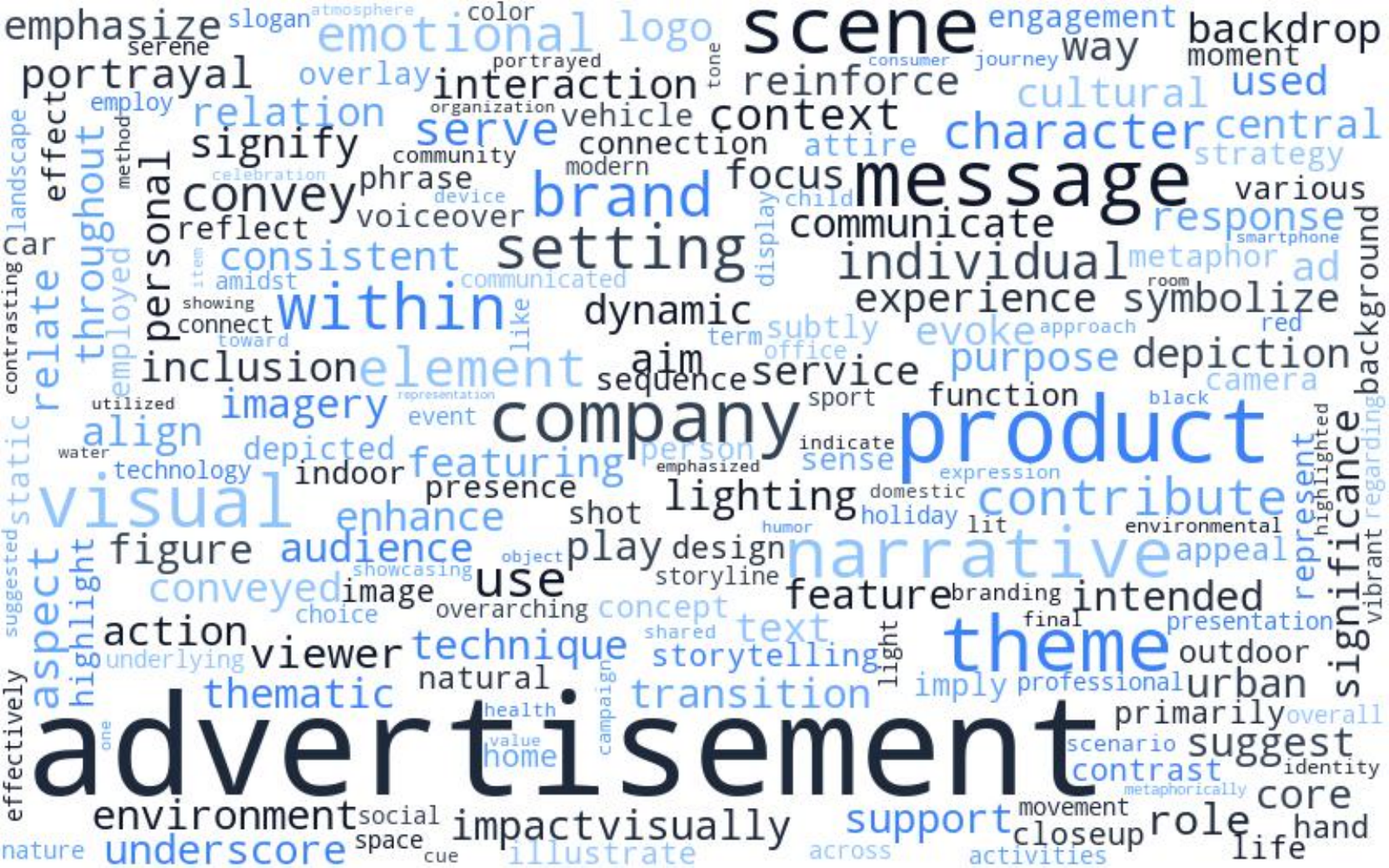}
      \caption{A word cloud of the ground truth answers of \adsqa~\cite{adsqa_2025_ICCV}, where the domain-specific words such as \textit{narrative}, \textit{engagement}, and \textit{storytelling} showcase the specialized reasoning in the advertising videos.} \label{fig:adsqa-word-cloud}
   \end{minipage}\label{fig_wc_ans}
\end{figure*}

\lens{} has three main features. 
\textbf{(1)}
Each image is annotated with rich text for all eight tasks. 
Thus, \lens{} supports to evaluate MLLMs to handle image-invariable prompts, from basic perception to compositional reasoning. 
\dcheck{This dataset provides the comprehensive evaluations that help the study of the
synergistic effects of different reasoning tasks.} 
\textbf{(2)} The images of \lens{} are collected from the social media, in which 53\% are published later than January 2025.
This feature ensures that when evaluating some recently released models, the influence of the models' inherent knowledge on the reasoning performance can be minimized as much as possible.
{\textbf{(3)}  
The diversity of object categories, scene types, multi-scale images, and bounding box annotations further facilitates numerous downstream tasks.
See a word cloud in Figure~\ref{fig:lens-word-cloud}.}

\paragraph{Difficulty.}

As reported in \cite{yao2025lens}, 
we evaluate \dcheck{20+} 
frontier MLLMs 
(\eg, Qwen2.5-VL-72B~\cite{bai2025qwen2}, InternVL3-78B~\cite{zhu2025internvl3}, Deepseek-VL2 \cite{wu2024deepseek}, Gemma3 \cite{team2025gemma}, GPT-4o~\cite{achiam2023gpt}, and Gemini2.5 Pro \cite{team2024gemini}) and two reasoning models (\ie, QVQ-72B-preview and Kimi-VL) that are released later than December
2024. None of them achieve an accuracy greater than 60\% in the reasoning tasks of \lens.  
Specifically, samples from the \lens{} are particularly challenging for the visual grounding task. This difficulty is attributed to a combination of factors, including complex and variable queries, a wide range of image resolutions, and a large number of small ground-truth bounding boxes, \etc.
As demonstrated in Table \ref{tab:rec_sota}, even Qwen2.5-VL-32B only achieves 48.47\% for \textit{Acc.@0.5}.

\subsubsection{\adsqa{} Dataset}

\paragraph{Statistics.}

\adsqa{} is an \underline{ad}verti\underline{s}ing video \underline{q}uestion \underline{a}nswering dataset sourced from 1,544 advertisement videos, providing 10,962 clips totaling 22.7 hours.
{Compared with image-based understanding tasks, VideoQA requires more computational resources.
To enable the participants to rapidly update their solutions, we thus sampled a subset from the \adsqa{} dataset to form the test set for Track \#3, consisting of 555 videos and over 3,000 questions.}
\adsqa{} comprises five open-ended QA tasks, each requiring distinct reasoning skills.
\begin{itemize}
\item Visual concept understanding: identifying and analyzing ad visuals.
\item Emotion recognition: detecting emotions and inferring character roles.
\item Theme and core message extraction: summarizing central theme and key messages.
\item Persuasion strategy mining: analyzing ad persuasion strategies.
\item Potential audience modeling: identifying and characterizing target audience.
\end{itemize}

\paragraph{Features.}

Ad videos are characterized by their domain-specific traits of being clue-rich and information-dense, owing to their inherent marketing logic, persuasive strategies, audience engagement, \etc.
Thus, \lens{} serves to assess the capability of MLLMs to perceive beyond the objective physical content of conventional visual domains.
Furthermore, understanding ad videos requires complex and specialized reasoning that is not compatible with the chain-style reasoning approach \texttt{``if A, then B''}.
\lens{} brings a novel domain that further extends the diversity of specialized reasoning in LLMs.
Figure \ref{fig:adsqa-word-cloud} illustrates the domain-specific words such as \textit{narrative}, \textit{engagement}, and \textit{storytelling}.

\dcheck{To our knowledge, \adsqa{} is the first video QA benchmark for advertisement domain, which is also the first ad benchmark for LLMs, 
with domain-unique features implicit, non-physical, mental, heuristic, \etc. 
It presents a new challenge to the current mainstream \texttt{``if A, then B''} LLM thinking approach, hence further extending the domain specialization reasoning scenarios for LLMs.
Moreover, it advances ad video understanding beyond physical-content dominated shallow perception towards deeper cognitive reasoning.}

\paragraph{Difficulty.}

Subtle visual details in ads pose a significant challenge for models, which often overlook them and fail to establish the connection between these elements, the video's theme, and the underlying creative intent.
We evaluated over 14 models, such as GPT-4o, Gemini 2.5 Pro, Qwen2.5-VL series, and LLaVA series~\cite{zhang2024video,li2024llava}, as reported in \cite{adsqa_2025_ICCV}. 
Our evaluations reveal that even the state-of-the-art Gemini 2.5 Pro attained a mere 60.7\% accuracy.

We have also conducted human evaluation by recruiting five non-expert evaluators who assessed the same randomly sampled subset (200 QA pairs). Each evaluator was required to answer the questions based solely on the provided videos, without any additional background information. The evaluators achieved an average accuracy of 71.4\%, demonstrating a clear gap between the state-of-the-art MLLMs and human performance.


\subsection{Competition Tracks}
\label{sec:tacks}

As stated in Section~\ref{sec:intro}, through the challenge of this year, we would like to make efforts to explore two challenging and understudied issues, \ie,  
\textbf{synergistic effects among reasoning tasks} and
\textbf{non-stepwise complex reasoning}.
Following this motivation, we define three competition tracks, where Track \#1 and Track \#2 share the same samples of \lens{} dataset to study the synergistic effects among reasoning tasks on the shared data distribution.
\adsqa{} dataset supports Track \#3 to evaluate non-stepwise complex reasoning in advertising.
In particular, the three tracks are defined as open-ended QA tasks.

\begin{itemize}
    \item {\bf Track \#1 Visual Grounding in Real-world Scenarios (VG-RS)} {evaluates the model’s scene perception,
object localization, and spatial reasoning abilities in complex scenarios.}
    \item {\bf Track \#2 Visual Question Answering with Spatial Awareness (VQA-SA)} {assesses how well the model performs spatial, commonsense, and counterfactual reasoning based on concrete physical content following user instructions.}
    \item {\bf Track \#3 Visual Reasoning in Creative Advertisement Videos (VR-Ads)} {probes the model’s
cognitive reasoning abilities in understanding implicit, non-physical, and abstract visual concepts in advertisement videos.}
\end{itemize}


\subsection{Evaluation Protocols}
\label{sec:protocols}

All our tracks take quantitative measures, and the higher score indicates the better outcome. The participants' submissions are evaluated on the open source platform \textit{EvalAI}. The evaluation protocols are as the follows.

Track \#1 VG-RS: 
To ensure the consistency with existing evaluation protocols, we follow the standard \textit{Acc.@0.5}  \cite{deng2021transvg,yao2024visual,dai2024simvg} to evaluate the detection performance for the visual grounding task. Specifically, 
given an input, 
the predicted bounding box is considered as correct only if the Intersection-over-Union (IoU) \cite{lin2014microsoft,zhang2022dino,yao2024ctod} between the prediction and the ground truth exceeds 0.5.

Track \#2 VQA-SA: 
We follow the prior work \cite{chen2024mllm} and employ a large language model \textit{GLM4-flash}\footnote{\url{https://github.com/zai-org/GLM-4}} as the automatic evaluator. 
For each predicted result and human annotation pair, our evaluation model generates multiple candidate responses and determines the correctness through majority voting. The final accuracy is calculated based on the number of correct predictions identified by the evaluator.

Track \#3 VR-Ads:
Following the recent practice~\cite{adsqa_2025_ICCV}, we employ \textit{gpt-4o-2024-08-06} to assist in evaluating text similarity between prediction and ground truth.
We provide each sample with the ground truth annotation and meta-information, which includes relevant details such as creative elements, storyline, and themes in the advertising video.
Moreover, we apply inclusion and exclusion rules to guide the scoring of text generated by the large language model. 
The inclusion rule requires the generated answer to incorporate as many elements of the ground truth  as possible. 
A score of 1.0 is awarded for a fully satisfactory answer, 0.5 for a partially matched one, and 0.0 otherwise.
The exclusion rule specifies that if the generated content includes elements that are not present in the ground truth and these elements fail to be inferred from the meta-information, then the content will be scored as 0.
We adopt the average score as the final evaluation metric.



\subsection{Terms, Phases, and Other Details}
\label{sec:rules_phases}

This challenge began on June 1st and ended on August 6th. 
Once the participants complete the registration, they will immediately obtain the test data for the competition tracks they are participating in and can start the submissions to test their solutions.
All competition tracks use the EvalAI evaluation server\footnote{\url{https://eval.ai/web/challenges/challenge-page/2552/overview}} to provide immediate feedbacks to the participants.
To offer participants the greatest freedom for innovation, the MARS2 2025 committee did not set any additional rules.
Both open-source and commercial models are allowed. Meanwhile, we have no restrictions on the size of the model. 

Based on the access rules for the leaderboard, it can be divided into two stages.

{\bf Stage \#1:} completely public {(01/06/2025 - 02/08/2025)}. To fully motivate the participants, the committee makes the real-time leaderboard public every day, including private submissions.
In this stage, we received $\sim$900 submissions.

{\bf Stage \#2:} priviate allowed (03/08/2025 - 06/08/2025). 
In the final stage, we discontinued the public leaderboard to increase competitive suspense and tension.
$\sim$300 submissions are received in Stage \#2.
Finally, we received over 1200 submissions.

\section{Methods and Results}
\label{sec:methods_results}

\paragraph{Overview.}

The MARS2 2025 Challenge has received widespread attention that
\dcheck{76} teams from the renowned academic and industrial institutions have registered and submitted \dcheck{1200+} entries in total.
The MARS2 committee has performed \dcheck{40+} state-of-the-art models as the baselines including both generalist MLLMs and task-oriented specialist models. 
After the submission deadline, \dcheck{40+} valid submissions have been included in our final ranking lists, in which a few teams contribute the commercial closed-source model based solutions.
The final scoring results are reported in Table~\ref{tab:final_leaderboard}. 
Finally, we have organized a large and comprehensive benchmark by combining our baselines and the participants' solutions, where the model sizes range from \dcheck{3B to 72B}. 
\dcheck{This benchmark provides the comprehensive comparison covering ``generalist models \vs specialist models'', ``large models \vs small models'', ``open-source models \vs closed-source models'', \etc.}

\begin{table*}[!ht]
  \centering
  \resizebox{1.0\textwidth}{!}{%
  \begin{tabular}{@{} r c c c c c  c c @{}}
    \toprule
    \multicolumn{8}{c}{{\bf Track \#1 Visual Grounding in Real-world Scenarios (VG-RS) on \lens{}}} \\
    Rank & Team & Method & Accuracy$\uparrow$ & Verified & Generalist  & Size &  Other Details \\
    \midrule
    1 & \textit{ActiveAlphaAgent} & Qwen2.5-VL-Distilled& 66.70& \checkmark & \checkmark & 7B& GRPO training, multi-turn tool usage\\
    2 & \textit{Star\_s} & Qwen2.5-VL \& G-Dino&64.83 &  \checkmark & -& 72B&  training-free, multi-scale input, prompt engineering\\
    3 & \textit{\tiny{Location depends on guessing}} & closed-source&64.30&  - & \checkmark& - & GRPO training, prompt engineering \\
    4 & \textit{SRCN-AIVL} & ensemble & 63.01& \checkmark  & \checkmark & -& ensemble methods, GRPO training\\
    5 & \textit{SUP} & ensemble & 61.90& \checkmark &- &- & ensemble methods, scene knowledge enhancement\\
    {BL} &  MARS2 & Qwen2.5-VL  & 48.47 & \checkmark    & \checkmark & 32B & state-of-the-art open-source model\\
    {BL} &  MARS2 & Qwen2.5-VL  & 46.94 & \checkmark    & \checkmark & 7B  & state-of-the-art open-source model\\
    {BL} &  MARS2 & Qwen2.5-VL  & 45.03 & \checkmark    & \checkmark & 3B  & state-of-the-art open-source model\\
    \bottomrule
  \end{tabular}
  }

    \resizebox{1.0\textwidth}{!}{%
  \begin{tabular}{@{} r c c c c c  c c @{}}
    \toprule
    \multicolumn{8}{c}{{\bf Track \#2 Visual Question Answering with Spatial
Awareness (VQA-SA) on \lens}} \\
    Rank & Team & Method & Accuracy$\uparrow$ & Verified & Generalist  & Size &  Other Details \\
    \midrule
    1 & \textit{Echoch} & ensemble& 79.03& \checkmark & \checkmark& -& ensemble methods, distillation, majority voting\\
    2 & \textit{Tele\_AI}& closed-source&72.60 &  - & \checkmark& - &  GRPO training\\
    3 & \textit{ActiveAlphaAgent}& Qwen2.5-VL-Distilled&69.72&  \checkmark& \checkmark& 7B & GRPO training, multi-turn tool usage\\
    4 & \textit{MILVLG\_HDU} & o3-2025-04-16 & 62.20&  \checkmark & \checkmark & -& prompt engineering\\
    5 & \textit{SRCN-AIVL} & ensemble & 54.60& \checkmark &\checkmark &- & ensemble methods, scene knowledge enhancement\\
    {BL} &  MARS2 & Qwen2.5-VL  & 54.10  & \checkmark    & \checkmark & 32B  & state-of-the-art open-source model\\
    {BL} &  MARS2 & GLM-4.1V-Thinking  & 51.32  & \checkmark    & \checkmark & 9B  & open-source reasoning model\\
    {BL} &  MARS2 & InternVL3  & 49.98  & \checkmark    & \checkmark & 38B  & state-of-the-art open-source model\\
    \bottomrule 
  \end{tabular}
  }

    \resizebox{1.0\textwidth}{!}{%
  \begin{tabular}{@{} r c c c c c  c c @{}}
    \toprule
    \multicolumn{8}{c}{{\bf Track \#3 Visual Reasoning in Creative Advertisement
Videos (VR-Ads) on \adsqa}} \\
    Rank & Team & Method & Accuracy$\uparrow$ & Verified & Generalist  & Size &  Other Details \\
    \midrule
    1 & \textit{gogogo\_truefaler} & Qwen2.5-VL & 56.35 & \checkmark & \checkmark& 72B &  feature engineering, CoT reasoning   \\
    2 & \textit{HNU-VPAI} & Qwen2.5-VL & 54.62 & \checkmark   & \checkmark & 72B & prompt engineering \\
    3 & \textit{ActiveAlphaAgent} & Qwen2.5-VL & 53.13 & \checkmark   & \checkmark & 72B & feature engineering, CoT reasoning  \\
    4 & \textit{rookiesllm} & Qwen2.5-VL  & 52.77 &  \checkmark  &  \checkmark & 72B &  feature engineering\\
    5 & \textit{mm618} & GLM-4.1V-Thinking& 52.74 & \checkmark & \checkmark & 9B & SFT, multi-hop reasoning  \\
    BL & MARS2 & Qwen2-VL& 41.41 & \checkmark    &  \checkmark & 7B & state-of-the-art open-source model  \\
    BL & MARS2 & MiniCPM-o & 43.23 & \checkmark & \checkmark & 7B & state-of-the-art open-source model \\
    BL & MARS2 & Qwen2.5-VL & 48.04 &  \checkmark  & \checkmark & 7B & ~~~~~~~~~~~~state-of-the-art open-source model~~~~~~~~~~~~ \\
    
    \bottomrule
  \end{tabular}
  }

  \caption{Final leaderboard of MARS2 2025. Only the Top5 teams of each track evaluated by the MARS2 committee are listed. See our website for full ranking lists. ``BL'': baseline methods implemented by the MARS2 committee. ``Generalist'': generalist model. ``Verified'': the method has been verified by the committee. \lens{} and \adsqa{} can be downloaded from our GitHub organization page \url{https://github.com/mars2workshop}.}
  \label{tab:final_leaderboard}
\end{table*}

\begin{table*}[!ht]
	\footnotesize
	\centering
	\resizebox{1.0\textwidth}{!}{%
		\begin{tabular}{l|c|c|ccccc|ccc}
			\toprule
			\multirow{2}[2]{*}{Method}  & Visual   & Linguistic  & \multicolumn{5}{c|}{Accuracy @ IoU} & \multicolumn{3}{c}{Scale-wise Accuracy}  \\
			 & Backbone & Backbone  &   $@0.5$  & $@0.6$ & $@0.7$  &   $@0.8$  & $@0.9$ & $ACC_{s}$  & $ACC_{m}$& $ACC_{l}$ \\
			\midrule    
			\multicolumn{11}{l}{\textbf{Methods based on predictive multimodal models: }}   \\ 
            \midrule    
			TransVG \cite{deng2021transvg}       & RN101 & BERT-B  & 8.73&7.57&6.29&4.40&1.69&0.01&2.01&23.64\\
VLTVG~\cite{yang2022improving}    & RN101 & BERT-B   & 11.04&9.60&7.75&5.33&1.99&0.00&2.80&29.70\\
CLIP-VG \cite{xiao2023clip}     & CLIP-B & CLIP-B    & 8.73 & 7.57 & 6.29& 4.40&1.69& 0.01 & 2.01 & 23.64\\ 
			MMCA \cite{yao2024visual}& RN101& BERT-B & 10.92&9.45&7.90&5.64&2.22&0.03 &2.79&29.31\\
    EEVG~\cite{chen2024efficient}    & ViT-B/16 & BERT-B  & 9.27 & 5.78 & 2.51 & 0.48 & 0.05&0.01 & 0.98 & 26.12 \\ 
            SimVG \cite{dai2024simvg}  & BEIT-3 & BEIT-3&16.46 &13.90 &11.12 &7.44 &2.70 &0.01 &3.10 &45.20\\
            G-DINO \cite{liu2025grounding}  & Swin-L & BERT-B   & 37.05 &33.92 &29.20 &22.57 &11.36 &24.87 &37.54 &52.32\\
			\midrule  
			\multicolumn{11}{l}{\textbf{Methods based on generative multimodal large language models: }}      \\ 
            \midrule    
            Groma-7B \cite{ma2024groma}  & DINOv2-L &  Vicuna  & 33.59& 29.95 &25.47 &18.73 &8.52 &11.58 &33.91 &58.59 \\
            Mova-7B \cite{zong2024mova}& Multi-expert& Vicuna &20.44 &13.10 &5.98 &1.09 &0.13 &5.06 &15.97 &40.36\\
            Ferret-7B \cite{you2024ferret} & CLIP-L & Vicuna & 23.26& 18.97 &13.95 &7.61 &1.84 &1.85 &19.49 &54.64\\
            Ferret-13B \cite{you2024ferret} & CLIP-L & Vicuna & 24.20 &19.81 &14.41 &8.12 &2.05 &2.26 &20.42 &56.31\\
            InternVL3-2B \citep{zhu2025internvl3}& InternViT-0.3B&Qwen2.5&7.89 &5.10 &2.85 &1.36 &0.33 &0.61 &3.46 &19.34\\
            InternVL3-8B \citep{zhu2025internvl3}& InternViT-0.3B&Qwen2.5&17.54 &13.23 &8.89 &4.94 &1.60 &3.23 &15.36 &35.21\\
            InternVL3-14B \citep{zhu2025internvl3} & InternViT-0.3B&Qwen2.5&29.53 &23.98 &17.25 &10.05 &3.00 &4.58 &27.80 &57.07\\
            InternVL3-38B \citep{zhu2025internvl3}&InternViT-6B&Qwen2.5&27.85&21.42 &15.00 &8.23 &2.37 &4.91 &25.81 &53.56\\
            LLM-wrapper \cite{cardiel2025llm} & Florence-2-L& Llama-3 &42.24 &38.78 &33.64 &26.10 &13.00 &21.99 &48.38 &58.39\\
            VLM-R1-3B \cite{shen2025vlm} & FE-ViT&Qwen2.5 &23.79 &19.91 &15.65 &10.53 &4.31 &8.15 &22.84 &40.94\\
		Qwen2.5-VL-3B \citep{bai2025qwen2} & FE-ViT&Qwen2.5&45.03&37.92&29.33&18.48&6.51 &29.14&50.54&57.15\\
            Qwen2.5-VL-7B \citep{bai2025qwen2} & FE-ViT&Qwen2.5& 46.94 & 39.39&29.94&18.38&6.26&28.87&54.12&60.24\\
            Qwen2.5-VL-32B \citep{bai2025qwen2} & FE-ViT&Qwen2.5& 48.47 & 40.66&30.78&19.15&6.63&30.93&55.04&61.48\\	
            \midrule  
			\multicolumn{11}{l}{\textbf{Methods in MARS2 competition: }}      \\ 
            \midrule  
            ActiveAlphaAgent &- &- & 66.71 &59.18 &48.07 &31.93 &12.28 &54.31 &68.07 &78.48\\
            Star\_s &- &- & 64.83 &57.74 &47.76 &34.43 &15.76 &49.86 &67.73 &78.05\\
            \tiny{Location depends on guessing} &- &- & 64.30 &56.02 &44.65 &29.76 &11.87 &50.99 &67.35 &75.57\\
            SRCN-AIVL &- &- &63.01 &56.83 &47.85 &35.14 &16.59 &47.57 &66.22 &76.37\\
            SUP &-&-& 61.90 &55.84 &47.53 &34.74 &15.97 &44.82 &64.79 &77.23\\
			\bottomrule
		\end{tabular}%
	}
    \caption{A comprehensive benchmark for MARS2 2025 Track \#1.}
	\label{tab:rec_sota}%
\end{table*}%

\begin{table*}[!ht]
	\footnotesize
	\centering
	\resizebox{1.0\textwidth}{!}{%
		\begin{tabular}{l|c|c|cccc}
			\toprule
			\multirow{2}[2]{*}{Method}  & Visual   & Linguistic  & \multicolumn{4}{c}{Accuracy @ Input Resolution} \\
			 & Backbone & Backbone  & 640 $\times$ 640 & 960  $\times$ 960& 1280  $\times$ 1280 &   1600  $\times$ 1600 \\
            \midrule    
            InternVL3-2B \citep{zhu2025internvl3}& InternViT-0.3B&Qwen2.5& 40.97&41.53&41.90&40.60\\
            InternVL3-9B \citep{zhu2025internvl3}& InternViT-0.3B&InternLM3-8B& 46.95& 46.54& 46.65&46.63\\
            InternVL3-14B \citep{zhu2025internvl3} & InternViT-0.3B&Qwen2.5& 50.28 & 51.15& 51.58&51.17\\
            InternVL3-38B \citep{zhu2025internvl3}&InternViT-6B&Qwen2.5& 50.88&50.97&49.98&51.08\\
		Qwen2.5-VL-3B \citep{bai2025qwen2} & FE-ViT&Qwen2.5&40.08&40.44&40.48&40.58\\
            Qwen2.5-VL-7B \citep{bai2025qwen2} & FE-ViT&Qwen2.5& 46.52 & 47.41&48.13&48.11\\
            Qwen2.5-VL-32B \citep{bai2025qwen2} & FE-ViT&Qwen2.5& 53.72 & 54.37&54.10 &54.09\\
            GLM-4.1V-Base-9B \citep{hong2025glm}& AlMv2-Huge&GLM-4-0414& 42.35 & 42.86&43.28 &43.73\\
            GLM-4.1V-Thinking-9B \citep{hong2025glm} & AlMv2-Huge&GLM-4-0414& 48.77 & 50.78&51.32 &51.13\\
			\bottomrule
		\end{tabular}%
	}
    \caption{Benchmark results  across varying input resolutions for MARS2 2025 Track \#2.}
	\label{tab:vqa_sota}%
\end{table*}%

\paragraph{Baselines.}

To offer the participants some reference and make our benchmark more comprehensive and diverse, the MARS2 committee provides \dcheck{40+} results as baselines, by using the state-of-the-art MLLMs such as Qwen2.5-VL and InternVL3.
In addition to generalist models, we further assessed more than 20 state-of-the-art specialist models proposed for Visual Grounding (VG) and Visual Question Answering (VQA). These models are included in our benchmarks, as summarized in Table~\ref{tab:rec_sota} and Table \ref{tab:vqa_sota}.

\paragraph{Observations and discussion.} The committee has carefully reviewed the solutions and code implementations. Our observations are summarized as follows.\\
\textbf{(1) Ideas.} Given the current prosperity of LLMs, some shared ideas are summarized from the solutions, which mainly involve ensemble, data augmentation, prompt engineering, alignment training, \etc.
To tackle complex reasoning tasks, most teams employ leading MLLMs as their base models and perform multi-step alignment through supervised fine-tuning (SFT) and reinforcement learning (RL).
Due to the recent success of the DeepSeek models, Group Relative
Policy Optimization (GRPO) \cite{shao2024deepseekmath} is widely used by the teams, \eg, \textit{ActiveAlphaAgent}, \textit{SRCN-AIVL}, and \textit{Tele\_AI}.\\
\textbf{(2) Data augmentation.} In the participants' solutions, domain-oriented data augmentation brings clear gains over the generalist base model (\eg, \textit{Location depends on guessing} Team), demonstrating the pattern novelty of our datasets \lens{} and \adsqa{} for the general-purpose models. \\
\textbf{(3) Reinforcement learning.} RL training works well for the complex and specialized reasoning as a kind of useful and promising alignment technique. The teams 
use the RL algorithm of GRPO for the post-SFT alignment training, achieving clear performance improvement.  \\ 
\textbf{(4) Prompt engineering.} Various prompts are designed carefully by the teams (\eg, \textit{Star\_s} and \textit{HNU-VPAI}), with a goal of using the detailed instructions to activate the reasoning of MLLMs. 
As evidenced by their prompt details, the samples of both \lens{} and \adsqa{} exhibit complex patterns and a high density of information.\\
\textbf{(5) Model collaboration.} The model collaboration is a promising direction, \eg, generalist and specialist collaboration, and this kind of solution achieves good performance in this challenge. 
For instance, the team \textit{Star\_s} employs Grounding DINO~\cite{liu2025grounding} as a task expert to verify and filter the bboxes generated by Qwen2.5-VL~\cite{bai2025qwen2}, earning second place in Track \#1.\\
\textbf{(6) Performance.} The results demonstrate that multimodal reasoning in complex scenarios and specialized domains is challenging, even using the strong LLMs as base models.
For example, it is observed that in Table~\ref{tab:final_leaderboard} the score of the winning solutions for \dcheck{VG-RS} task does not exceed \dcheck{70\%}.
In Track \#3, the best accuracy (56\%) still has a clear gap if compared with the human performance ($\sim$70\% \cite{adsqa_2025_ICCV}).
These results highlight the challenging nature of reasoning on the \lens{} and \adsqa{} datasets.

The top-performing methods for each track are described in the following parts. 
For further discussion, please refer to Section~\ref{sec:discussion_outlook}.


\subsection{Track \#1 Visual Grounding in Real-world Scenarios (VG-RS)}

\subsubsection{VG-SMART}
\label{sec:VG-SMART}

(submitted by \textit{ActiveAlphaAgent} Team)

The champion solution of Track \#1 is a multi-stage post-training 
\underline{V}isual \underline{G}rounding
method: \underline{S}NR-Driven Data Synthesis based \underline{M}ulti-Stage \underline{A}lignment Combining Supervised Fine-Tuning
and \underline{R}einforcement \underline{T}raining, termed VG-SMART. 
This method begins with synthesizing a high-quality dataset via ensembling the state-of-the-art models and using a filtering process based on a custom signal-to-noise ratio (SNR) metric. The key idea of this method is the multi-stage training process built upon the Qwen2.5-VL-72B-Instruct model~\cite{bai2025qwen2},
which involves grounding-based supervised fine-tuning (SFT) followed by a reinforcement learning (RL) stage with an IoU-centric reward function. 
The final step is distilling the knowledge from 72B model to Qwen2.5-VL-7B to optimize inference speed.

\paragraph{Data synthesis and filtering.} The data augmentation mainly includes two steps:

\noindent {\bf (1) Synthesis of domain-specific data.}
They perform a domain-based classification of the Track \#1's test set, establishing four main categories.
\begin{itemize}
    \item {Transportation hubs:}  $\sim$20.0\%
    \item {Urban \& life scenes:} $\sim$21.0\%
    \item {Indoor spaces:} $\sim$44\%
    \item {Educational \& public institutions:} $\sim$15\%
\end{itemize}

{In addition to visual data analysis, a fine-grained examination of the questions is conducted to understand the types of reasoning demanded by the test set. The queries are analyzed from multiple perspectives, including:  object categories, object attributes, and spatial relationships. The results reveal that each domain poses distinct grounding challenges. For example, transportation and urban scenes frequently exhibit occlusion or contain minuscule objects.}

To capture these complex patterns, the team synthesizes a dataset using an ensemble of the state-of-the-art MLLMs \cite{bai2025qwen2, coreteam2025mimovltechnicalreport, zhu2025internvl3}. 
The approach leverages multi-model voting to identify high-confidence responses and employs difference analysis to create challenging and diverse samples.
This process yields an initial dataset of  $\sim$11k samples.

\noindent {\bf (2) Data filtering by signal-to-noise ratio and pass rate.}

The synthesized dataset encompasses samples of varying difficulty (simple, moderate, and difficult) but may contain noise. To enhance its quality, a signal-to-noise ratio (SNR) metric is defined as $\text{Accuracy}_{\text{synthetic}} / \text{Accuracy}_{\text{base}}$ and employed for filtering. The accuracy values are derived from the committee's online evaluation system. Only samples with an SNR greater than 1 are retained.

To quantify the difficulty distribution of the refined samples, the team uses the Qwen2.5-VL-7B-Instruct~\cite{bai2025qwen2} with a temperature setting of 1.0 to run eight inference passes per sample. The pass rate@8 (the number of correct predictions out of eight attempts) serves as the metric to categorize each sample as easy, medium, or hard. 
As a result, the difficulty distribution of the refined samples becomes balanced, forming a high-quality training set that provides more comprehensive learning signals for the subsequent stages of SFT and RL.

\paragraph{Supervised fine-tuning.}

The first stage is a cold-start~\cite{guo2025deepseekr1} supervised fine-tuning phase. The primary goal of this stage is to adapt the Qwen-VL-72B-Instruct model to the specific domains, question styles, and output formats of this competition track, meanwhile establishing a robust baseline for the reinforcement learning phase. 
Thus, one necessary step is aligning the training samples with the Qwen-VL instruction format. 
For image preprocessing, they use the standard \textit{Qwen-VL-Utils}, setting \textit{max\_pixels} to $1280 \times 28 \times 28$ and applying a \textit{smart\_resize} transformation to both the images and their corresponding ground truth bounding boxes.
The team employs a full-parameter fine-tuning strategy for both the adapter and the large language model, while the ViT~\cite{dosovitskiy2020image} backbone is frozen.

\paragraph{Reinforcement learning.}

Motivated by \cite{guo2025seed1, yue2025does}, the team deliberately selects only the medium and hard samples from their training set, as identified by the \textit{pass\_rate@8}-based analysis. 
They also use a modified IoU reward function aimed at amplifying the feedback signal around the critical Intersection over Union (IoU) decision boundary of IoU=0.5. For predictions with an IoU greater than 0.5, a non-linear function is applied to attenuate the reward difference, preventing the model from becoming overly conservative on already correct samples. Meanwhile, for predictions in the range of 0.3-0.5, the reward function magnifies the negative feedback. 
This approach encourages more aggressive learning from critical errors and significantly improves fine-grained decision-making.

This team uses \textit{verl} framework \cite{sheng2024hybridflow} 
and Group Relative Policy Optimization (GRPO)
\cite{shao2024deepseekmath} with a rollout parameter of 8. 
The model is trained for three epochs. 
For validation during training, they use a custom metric defined as the pass rate for predictions achieving an IoU greater than 0.5.

In addition to the aforementioned approach, this team reports that they explore a thinking-based agent \cite{zheng2025deepeyes} that uses multi-turn tools, such as zoom-in and grid overlays.

\paragraph{Distillation.}

To balance performance and efficiency, this team creates a final model by distilling the knowledge from Qwen2.5-VL-72B-Instruct into Qwen2.5-VL-7B-Instruct \cite{zhang2025hkd4vlm, zhang2025tokenfocus}. 
Team \textit{ActiveAlphaAgent} secured the first place on the Track \#1's final leaderboard, with a score of 0.6671.

\subsubsection{DCM-VG}

(submitted by \textit{Star\_s} Team)

This team proposes a solution, \underline{D}etector \underline{C}onsensus Guided \underline{M}ulti-query \underline{V}isual \underline{G}rounding (DCM-VG).
The motivation is to combine the generalist model and the specialist model.
Based on tailored prompting strategy, the generalist model Qwen2.5-VL~\cite{bai2025qwen2} generates bounding box candidates, which are then verified and refined by the task-specific model Grounding DINO~\cite{liu2025grounding} to select the optimal result.

\paragraph{Prompting strategy.}
The team implements the following prompt to consolidate all detection targets within an image into a single prompt, which is then fed into Qwen2.5-VL to stimulate the model’s ability to leverage inter-object associations. 
\begin{quote}
    \textit{I will provide you with an image and several questions. \\
……\\
Return the accurate 2D bounding box coordinates for all main objects mentioned in the questions. \\
……\\
If an object mentioned in a question is not found in the image, set its ``bbox\_2d'' value to [0, 0, 0, 0].}
\end{quote}

Despite using a multi-target prompt, Qwen2.5-VL still regularly fails to identify small or obscured objects. Therefore, the team proposes an enhanced prompt for such failure cases.
The enhanced prompt template is presented below.
\begin{quote}
    \textit{I will provide you with an image and several questions. \\
You need to note that these targets are usually obscured or small. You must select objects that strictly meet the textual description.\\
……\\
Return the accurate 2D bounding box coordinates for all main objects mentioned in the questions.\\ 
……\\
You must provide the most likely one ``bbox\_2d''.}
\end{quote}

Moreover, Qwen2.5-VL is sensitive to variations in image scale. To improve its robustness, the team employs a multi-resolution inference strategy, processing each image at multiple resolutions. The task expert model then selects the most confident results, leading to more consistent and reliable grounding performance.

\paragraph{Selection strategy.}
The team employs Grounding DINO as a task expert to verify and filter the candidate boxes generated by Qwen2.5-VL, a solution which earned second place in Track \#1 with a final score of 0.65.

\subsubsection{VR-VG}

(submitted by \textit{Location depends on guessing} Team)

This team implements a multi-stage alignment solution involving data augmentation, SFT, and RL.
The main idea is to design a 
\underline{v}erifiable \underline{r}eward for \underline{v}isual \underline{g}rounding (VR-VG).
To alleviate data scarcity, a five-step automated pipeline is proposed, including global captioning, entity extraction, instance grounding, region refining, and question synthesis.
Leveraging the strong tools such as Qwen-2.5-VL-72B and Grounding DINO, 
the team generates a large-scale synthetic dataset comprising 2.7 million diverse question-answer pairs.

\paragraph{RL with GRPO and verifiable rewards.}

The proposed model is designed to jointly leverage fine-grained local details (high-resolution) and global scene context (low-resolution) via the lightweight cross-attention.  
Following the practice of the two-stage training strategy, they first perform supervised fine-tuning on the synthetic dataset, and then apply reinforcement learning with verifiable rewards to further enhance grounding accuracy.
The reward is a weighted sum of $R_{fmt}$ and $R_{{iou}}$ defined as follows.

\noindent   \textbf{(1) Verifiable format reward $R_{{fmt}}$}: 
a binary validation signal that verifies whether the output strictly adheres to the specified schema:
  \textit{``\textless think\textgreater...\textless/think\textgreater\textless answer\textgreater(x1, y1), (x2, y2)\textless/answer\textgreater''},
  ensuring both correct tag structure and proper coordinate tuple format.

\noindent  \textbf{(2) Dense IoU reward $R_{iou}$}: 
a continuous score that offers fine-grained spatial supervision proportional to the IoU between the predicted and ground truth boxes.

This team reports that this combination of rewards improves both semantic correctness and spatial precision, producing an additional +3.2 mIoU.
More details can be found in their code implementation.

\subsubsection{RVG-CVME} 

(submitted by \textit{SRCN-AIVL} Team)

This solution, FVG-CVME, is a \underline{r}einforced 
\underline{v}isual-spatial 
\underline{g}rounding model that integrates 
\underline{c}andidate 
\underline{v}oting and 
\underline{m}odel 
\underline{e}nsembling. It employs a two-stage training strategy (SFT + RL) combined with training-free ensemble techniques.

\paragraph{Two-stage training.}
Before training-free ensembling, the team optimizes a 3B base model by a two-stage training of SFT and RL.

\noindent \textbf{(1) Cold-start supervised fine-tuning.}
The team observes that \lens{} dataset contains diverse scenarios like in/outdoor, city street, shopping mall, \etc, in which there are more than one-third referring cases are person-related. Thus, this team uses HumanRef-CoT~\cite{jiang2025rex} as their SFT dataset. 
HumanRef-CoT is a large-scale human-centric referring expression dataset designed for multi-instance human referring in natural scenes.
The base model selected for this stage is Qwen2.5-VL-3B, with its vision encoder and connector frozen during SFT.

\noindent \textbf{(2) Reinforcement learning training.}
The team adopts the GRPO algorithm to conduct reinforcement learning training, aiming to boost the model's performance. Prior to this, Qwen2.5 is utilized to extract all entity categories.
For example, given a referring expression ``relatively the middle socket'', its entity category will be extracted as ``socket''. 
Then, Grounding DINO extracts all candidate boxes related to ``socket'' in the image. 
The target is to drive the model to select the bbox to match the description.
An option for the reward function is F1 score. 
Moreover, this team also expects that the model has zero-shot regression ability when the given candidate boxes fail to match the ground truth.
Thus, they additionally add a bbox IoU reward function working in conjunction with the F1 score.

\paragraph{Ensemble.}
This team employs VLM-R1-3B~\cite{shen2025vlm}, Qwen2.5-VL-32B, and Qwen2.5-VL-72B as their base models for result aggregation.
They then apply weighted boxes fusion (WBF)~\cite{solovyev2021weighted} incorporating two domain-specific experts (OV-DINO~\cite{wang2024ov} and LLMDet~\cite{fu2025llmdet}) along with a reasoning model (GLM-4.1V-Thinking~\cite{hong2025glm}) as an assistant model to address unprocessed parts.

\subsubsection{KEME}

(submitted by \textit{SUP} Team)

The team proposes a method named KEME (\underline{k}nowledge \underline{e}nhancement with \underline{m}ultistep \underline{e}nsemble), which consists of a three-stage process.

\paragraph{Input preprocessing.} 
Both visual and textual modalities undergo preprocessing.
To address the inherent resolution limitations of vision transformers, they employ two distinct resizing strategies:
(a) resizing all images to a fixed resolution of 1120×1120 pixels, and
(b) resizing the longest edge of each input image to 1400 pixels to preserve the original aspect ratio and padding the shorter edge.
To reduce semantic ambiguity in queries, the team employs a bilingual prompting strategy. Specifically, each question is translated into Chinese, and both the original English version and its Chinese translation are provided to the model. This dual-language approach improves the robustness to phrasing variations and mitigates misgrounding caused by ambiguous expressions.

\paragraph{Contextual reasoning enhancement.}

For each input image, Qwen2.5-VL-72B~\cite{bai2025qwen2} is prompted to generate rich contextual information, including a scene category, a global caption, and a description of spatial relation. For every image–question pair, the model identifies the key subjects mentioned in the question, which are subsequently grounded to the corresponding regions using Florence-2~\cite{xiao2024florence}. 
This assembled context including scene tag, caption, and relevant bounding boxes is processed by Qwen2.5-VL under varied prompting strategies to produce multiple outputs for subsequent stages.

\paragraph{Model ensemble.}
This method employs Qwen2.5-VL as an evaluator to assess detector candidates such as Florence-2, Grounding DINO, and LLMDet. 
For each query, Qwen2.5-VL evaluates the proposed bounding boxes and selects the best-matching candidate. If no suitable candidate is identified, a deterministic fallback strategy is applied.

\subsubsection{A VLM-R1-based Visual Grounding Method}

(submitted by \textit{CV and RL} Team)

This team participated in Track \#1, achieving 8th place.
They use Qwen2.5-VL-7B~\cite{bai2025qwen2} and RefCOCOg~\cite{yu2016modeling} as the base model and the training set, respectively. 
Within the VLM-R1~\cite{shen2025vlm} framework, they implement the GRPO algorithm~\cite{shao2024deepseekmath} by designing a custom reward function.
The model's output is passed through a regularization function to extract bounding box coordinates. The reward is determined by the Intersection over Union (IoU) between the predicted bounding box and the ground truth: a reward of 1 is given if IoU $\geq$ 0.5, and 0 otherwise.


\subsection{Track \#2 Visual Question Answering with Spatial Awareness (VQA-SA)}

\subsubsection{RSVT}

(submitted by \textit{Echoch} Team)

The champion team of this track introduces a two-stage fine-tuning method, \underline{r}ejection \underline{s}ampling and \underline{v}iewpoint \underline{t}ransformation, termed RSVT. 
In the first stage, a bilingual cold-start corpus is built using multimodal benchmarks and high-quality translations for linguistic consistency. 
In the second stage, self-consistency generation and cross-view augmentation enrich multi-view samples, improving spatial reasoning generalization.

\paragraph{Data preparation.}

The data preparation is divided into two stages, drawing from both publicly available datasets and synthetic data sources to ensure diversity.

\noindent \textbf{(1) First stage.} Multiple multimodal spatial reasoning benchmarks (\eg, 3DSRBench \cite{ma20243dsrbench}, DORI-Benchmark \cite{nichols2025right}, OpenSpaces \cite{spacethinker2025}, 
SpaceOm \cite{spaceom2025}, Spatial\_MM\_CoT \cite{shiri2024empirical}, SpatialMQA \cite{liu2025can}, and ViewSpatial-Bench \cite{li2025viewspatial}) are adopted to broaden the sample and task diversity. 
They use GPT-4o for high-quality translation and rewriting to ensure linguistic consistency.
In addition, the VQASynth \cite{spacethinker2025} toolkit generates around 10k synthetic spatial-relation QA pairs, which are combined with benchmark datasets to yield about 170k samples as the initial training set.

\noindent \textbf{(2) Second stage.} Advanced data augmentation is applied to improve diversity. 
Multiple MLLMs (\eg, OpenAI o3 \cite{openai_o3_blog_2025} and Gemini 2.5 Pro \cite{comanici2025gemini}) are used to incorporate scene and viewpoint cues. 
A self-consistency strategy then generates multiple QA candidates for each sample and selects the most representative ones.
Further expansion is achieved with open-source models (\textit{e.g.}, InternVL3-38B \cite{zhu2025internvl3} and Qwen2.5-VL-32B \cite{bai2025qwen2}), alongside viewpoint transformation augmentation to enhance generalization in VQA-SA reasoning.

\paragraph{Training details.}

This team trains three MLLMs, \ie, InternVL3-14B, InternVL3-38B, and Qwen2.5-VL-32B.

\noindent \textbf{(1) First stage.} For cold start, the team fine-tunes the InternVL3-14B model on 170k bilingual samples using low-rank adaptation (LoRA) \cite{hu2022lora}. The model is trained at a resolution of 448×448 to capture essential spatial features. Owing to constraints in model size and training time, they train InternVL3-38B and Qwen2.5-VL-32B exclusively on the Chinese samples for only three epochs.

\noindent \textbf{(2) Second stage.} 
The three models undergo additional training for five epochs using a second-stage dataset, which is augmented via viewpoint transformation and self-consistency strategies. 
In this stage, the image resolution is enhanced to 1024×1024 pixels to capture more fine-grained spatial details.

\paragraph{Reasoning enhancement.}

Each model independently generates predictions, and the final output is determined via majority voting among the three models.
To improve the recall of Chinese spatial expressions, post-processing heuristics are applied, for instance, mapping ``front-right" to ``right-front" or ``front-right side".
For multi-turn dialogue scenarios, they append each subsequent question under the same image with the camera perspective established in the first turn.
This approach ensures that the viewpoint remains consistent and unambiguous throughout the interaction.

\paragraph{Tricks.}
The committee has observed that this team employs two effective strategies. First, they provide deliberately ambiguous responses, such as offering multiple possible answers to location-based questions. This increases the probability of matching the ground truth. Second, they present the same answer in various grammatical forms or phrasings, reducing false negatives in large-scale automated evaluation.

\subsubsection{SMART}

(submitted by \textit{Tele\_AI} Team)

The team introduces the SMART framework, comprising modules for \underline{s}ynthesis, \underline{m}odel \underline{a}daptation, \underline{r}easoning, and \underline{t}esting. 
They construct a {m}ultimodal {r}easoning and {t}hinking {d}ataset (MRTD) and employ a two-stage training pipeline. 
In the final evaluation, the team secured second place with an accuracy score of 0.7261.

\paragraph{Training data synthesis.}

The team designs an automated workflow for generating diverse spatial reasoning training samples as follows.

\noindent \textbf{(1) Spatial scene rendering.} 
The team leverages Blender's programmable API to build multi-object scenes with predefined spatial relationships. 
An automated filtering process removes invalid cases, while collision and occlusion checks ensure visibility. 
The 3D coordinates and orientations of all objects are then recorded for subsequent use.

\noindent \textbf{(2) QA pair generation.} 
The team constructs question templates across three difficulty levels, embedding object attributes such as category, color, and spatial relations. 
Answers are derived through geometric analysis of recorded object coordinates to capture relations like ``left of", ``front–back", or ``behind".
An example of the hard question is illustrated below.

\begin{quote}
    \textit{In the picture, in which direction is the sky-blue road bicycle in relation to the red car? Please choose from front, back, left, right, above, below, front left, front right, left behind, right behind.}
\end{quote}

\paragraph{Model fine-tuning.}
The team implements a two-stage fine-tuning pipeline combining supervised learning and reinforcement learning.

\noindent \textbf{(1) Location-aware data preparation.}
The team employs Qwen2.5-VL-32B \cite{bai2025qwen2} to generate QA samples containing explicitly position-aware reasoning chains. 
The synthetic data is integrated with the GRiD-3D \cite{lee2022right} and GQA \cite{hudson2019gqa} datasets in a ratio of 2:1:1 and augmented with original QA pairs to enhance dataset diversity. 
An example of location-aware reasoning is provided below.
\begin{quote}
    \textit{Question: ``In which direction is the sky-blue road bicycle in relation to the red car? ''} \\ 
    \textit{\textless think\textgreater The sky-blue bike is positioned at the bottom left of the image, ... To determine the direction of the blue bike relative to the car, we need to evaluate its location from the car's view. From the car's viewpoint, the sky-blue bike is in front right of it. \textless /think\textgreater}
\end{quote}

\noindent \textbf{(2) Cold-start training and reinforcement learning.}
In the first stage, cold-start training strategy \cite{guo2025deepseekr1} is performed on their MRTD dataset, aiming to minimize the average negative log-likelihood over reasoning tokens and answer tokens.
In the second stage, reinforcement learning is performed using the Decoupled
Clip and Dynamic sAmpling Policy Optimization (DAPO) \cite{yu2025dapo}, in which reward signals are computed via normalized advantage functions to ensure stable optimization guidance.

\paragraph{Inference strategy.}
The team's reasoning strategy centers on spatial relationship modeling and contextual dependency. 
They leverage GPT-4o \cite{achiam2023gpt} to refine context-dependent questions, enhance image resolution to retain fine-grained spatial cues, and generate multiple reasoning trajectories. 
A process-level reward model, VisualPRM-8B \cite{wang2025visualprm}, evaluates these trajectories to select the answer most consistent with semantic logic, thereby improving stability and reliability in spatial reasoning.

\subsubsection{STAGES}

(submitted by \textit{ActiveAlphaAgent} Team)

Based on the analysis of the Track \#2 dataset, the team proposes the \underline{s}ynthetic \underline{t}raining and \underline{a}daptive \underline{g}raduated \underline{e}nhancement \underline{s}trategy (STAGES). 
The method first optimizes the prompts and then adopts a three-stage training strategy that involves large-scale SFT on synthetic data, in-domain SFT on high-confidence samples, and knowledge distillation for model efficiency.

\paragraph{Prompt optimization.}
Based on a preliminary analysis of the Track \#2 dataset, the team optimizes prompts before 
training. 
A key strategy is to enrich the input prompt for a given question with the complete list of all VQA-SA and VG-RS questions that share the same image.
This method results in an accuracy improvement of 2–7 percentage points in multiple models, with stronger reasoning models exhibiting greater gains. 
Further performance gains are achieved by customizing the prompts according to model type, image category, and question type.

\paragraph{Three-stage training strategy.}
With the optimized prompts established, STAGES proceeds with a three-stage training strategy, preparing a distinct dataset for each stage.

\noindent \textbf{(1) Large-scale task-oriented SFT.}
This team performs large-scale supervised fine-tuning \cite{yue2025does} using synthetic data generated by Qwen2.5-VL-72B-Instruct \cite{bai2025qwen2}.
The data construction process begins by sourcing images from public datasets and the internet, then filters them for visual similarity to the Track \#2 samples, yielding approximately 150k valid images.
Based on these images and MLLMs, 1.5 million new questions are generated, for which
the team uses their optimized prompts to generate answers.
Then a strict filtering approach of ensemble consistency and model scoring is used, producing 800k high-quality QA pairs.

\noindent \textbf{(2) In-domain SFT.}
In this stage, to align with the distribution of the Track \#2 dataset, optimized prompts are applied across multiple MLLMs to generate answers for the Track \#2's 6.5k VQA-SA questions.  
An ensemble and model-based scoring strategy yields a dateset of high-confidence samples with an accuracy of 79.47\%.
The stage-one model is then fine-tuned on these in-domain samples, leading to further performance gains.

\noindent \textbf{(3) Knowledge distillation.}
To reduce the inference cost of the 72B model, the final stage applies knowledge distillation \cite{zhang2025hkd4vlm,zhang2025tokenfocus}, transferring knowledge from Qwen2.5-VL 72B to a smaller Qwen2.5-VL 7B model. 
Notably, the 7B student achieves a precision of 69.72\% in VQA-SA, matching the performance of its 72B teacher.

\subsubsection{o3-MQP}

(submitted by \textit{MILVLG\_HDU} Team)

The team develops a \underline{m}ulti-\underline{q}uery \underline{p}ipeline based on OpenAI o3 (o3-MQP) \cite{openai_o3_blog_2025} that 
is designed to deeply analyze fine-grained spatial relationships between objects within an image and to execute complex visual reasoning. 
This solution comprises two primary stages.

\paragraph{Preprocessing stage.}
The team observes that in the Track \#2's dataset, questions associated with the same image often exhibit logical dependencies and semantic correlations. 
Consequently, during the preprocessing stage, they perform difficulty-based grouping of questions for each image. 
This approach prioritizes addressing simpler and more fundamental questions first, thus supplying contextual cues for answering more complex questions in subsequent steps.

\paragraph{Prompting stage.}
During the prompting stage, the team designs a multi-query prompting template to query the OpenAI o3 model. 
The template batches multiple questions about the same image into a single prompt, and the model returns answers to all queries in a single-pass response. 
Coupled with o3's tool-augmented reasoning, the template enables calls to external tools, such as bounding-box annotation, image cropping, and object classification, to tackle complex visual reasoning. 
By leveraging shared cross-question context within the prompt, the approach improves the consistency and accuracy of the multi-query answers.

\subsubsection{CCP4SR}

(submitted by \textit{SRCN-AIVL} Team)

The team proposes a method of \underline{c}oreference and \underline{c}ausality \underline{p}rompting for \underline{s}patial \underline{r}easoning (CCP4SR), which enhances spatial reasoning in multi-turn VQA-SA tasks by integrating coreference resolution, causal prompting, and deep learning assisted spatial reasoning. 
This approach is designed to mitigate ambiguities from omitted referents and to improve the model’s ability to interpret directional relationships between entities.

\paragraph{Multi-turn coreference resolution with causal prompting.}
After reviewing the test set, the team observes that answering questions individually introduces coreference ambiguity, 
and prevents the effective use of causal relationships across consecutive questions.
When isolated, these questions often omit explicit subject or object descriptions, making it challenging for MLLMs to recover the missing information from the query alone.
Such omissions include not only the reference subject but also the perspective and answer options established in earlier rounds. 
Simply concatenating the QA pairs from previous questions as input for the current one further risks ambiguity, as the coreference chain becomes unstable with increasing context length.
To address this issue, the team designs a preprocessing step that treats each image as a unit encompassing all its associated questions while preserving their original order. 
This process aims to capture as many inter-referential dependencies and causal connections across multiple rounds as possible.

\paragraph{Depth-augmented spatial reasoning.}
The team incorporates depth information into spatial reasoning to improve performance on direction-related questions. 
Drawing on prior research in VQA spatial reasoning, they select Ovis2-34B~\cite{lu2024ovis} as the primary model, balancing parameter scale and computational constraints.

For the SpaceOm dataset \cite{chen2024spatialvlm}, despite being fine-tuned on synthetic spatial reasoning data, its low training resolution and small model size (3B) limit its ability to identify small entities and fully comprehend questions. 
Larger models (32B, 34B) demonstrate improved question understanding but still exhibit frequent errors on direction-related tasks requiring precise attribute determination.
To address this issue, the team adapts the SpatialRGPT \cite{cheng2024spatialrgpt} pipeline. 
Direction-related questions are first filtered from the test set, and relevant entities and reference objects are extracted using an LLM. 
Depth maps from Depth Pro \cite{bochkovskii2025depthprosharpmonocular} are then incorporated to provide geometric cues, guiding the model towards depth-informed reasoning processes.

However, in practice, models without task-specific training struggle to leverage bounding boxes and depth for complex reference transformations through prompt engineering. 
While this approach helps address basic depth-related questions, its overall performance remains slightly inferior.

\subsubsection{Prompt4SA}

(submitted by \textit{Tang\_TUTE} Team)

The team (ranked \#8) develops the Prompt4SA system to address spatial reasoning challenges such as coreference ambiguity, compositional attributes, occlusion, and viewpoint variation. 
Built on the MS-SWIFT LLM \cite{zhao2025swift} with vLLM~\cite{kwon2023efficient} as the inference backend, it employs a cognition-guided structured prompting strategy that decomposes reasoning into five stages. 
By defining the reference frame before computing relations, the method reduces answer drift and improves consistency, achieving high accuracy and stable performance in the official evaluation of Track \#2 without task-specific fine-tuning.

\paragraph{Structured prompt for spatial reasoning.}
This team introduces a structured prompt strategy to stabilize spatial reasoning and reduce ambiguity through a five-stage process. 
It begins by classifying the question to determine cognitive complexity, then performs feature binding and object recognition to resolve composite descriptors and anchor objects. 
Next, it establishes the proper reference frame, objective, subjective, or ego-centric, before computing spatial relations such as directions, distances, and occlusions. 
Finally, the output is verified for consistency and formatted as XML-style tags to support reliable post-processing.

\paragraph{Training and inference details.}
This approach is implemented on top of the {VllmEngine} in MS-SWIFT, employing MLLMs such as InternVL3 \cite{zhu2025internvl3} as the base model. 
During post-processing, the final predictions are extracted from the
\textit{\textless answer\textgreater...\textless/answer\textgreater}
tags in the model output, and all QA pairs for each image are aggregated into a single visualization panel. 


\subsection{Track \#3 Visual Reasoning in Creative Advertisement Videos (VR-Ads)}

\subsubsection{Hi-CoT}
\label{sec:Hi-CoT}

(submitted by \textit{gogogo\_truefaler} Team)

The champion of this track proposes a \underline{hi}erarchical \underline{c}hain-\underline{o}f-\underline{t}hought based method for VR-Ads. Their idea is to guide a multimodal large language model to perform intention-driven reasoning through structured prompts, emulating human cognitive processes that progress from perceptive understanding to higher-order reasoning. 

\paragraph{Method.}
The pipeline comprises four sequential steps:

\noindent \textbf{(1) Global understanding via audio-visual integration.}
The audio in the video is transcribed by automatic {s}peech {r}ecognition tools~\cite{ffmpeg,radford2023whisper}. A custom-designed prompt is then employed to guide the MLLM in conducting a global analysis of the video. This initial step aims to extract high-level semantic information, including the advertisement's core theme, emotional tone, persuasive intent, and overall narrative structure.

\noindent \textbf{(2) Segment-level reasoning with contextual alignment.}
After extracting the global context, a fine-grained shot-level analysis (segmented by scene changes) is conducted to address detailed questions.
The video is partitioned into discrete shots based on scene transitions, and for each segment, the MLLM generates time-stamped visual-semantic descriptions.
Subsequently, question-driven prompts are employed to align user queries with relevant segments, improving query-vision alignment.

\noindent \textbf{(3) Hierarchical reasoning over global and local observations.}
With key observations identified, the model proceeds to perform higher-level reasoning. This stage involves causal and commonsense inferences based on evidence collected in the previous steps. The model connects the ``what'' (observations) with the ``why'' (underlying reasons), bridging the gap between perception and cognition.

\noindent \textbf{(4) Answer generation.}
Finally, the model integrates all intermediate reasoning steps, including global context, localized observations, and causal inferences, to generate the final answer. This structured approach ensures that the output is not merely a direct retrieval of facts but a well-reasoned conclusion grounded in a logical chain of thought.

\paragraph{Implementation details.}
The Qwen2.5-VL-72B~\cite{bai2025qwen2} serves as the base model. 
A custom preprocessing pipeline is developed to enhance video understanding performance and address the data scarcity. 
The audio processing stage utilizes FFmpeg v5.1 for audio stream extraction, with subsequent transcription performed by the Whisper Large-v3 model~\cite{radford2023whisper} to facilitate multimodal analysis. 
The visual processing pipeline performs temporal segmentation using PySceneDetect v0.6.1~\cite{pyscenedetect}, with a scene change threshold of 30. 
Each segmented video shot then undergoes detailed visual description generation via Qwen-VL, using well-designed prompting strategies.

\subsubsection{TFA}

(submitted by \textit{HNU-VPAI} Team)

This team's solution adopts a \underline{t}raining-\underline{f}ree \underline{a}pproach (TFA). 
Their experiments are conducted on Qwen2.5-Omni~\cite{xu2025qwen2}, Qwen2.5-VL~\cite{bai2025qwen2}, and GPT-4o~\cite{achiam2023gpt}, and results indicate that Qwen2.5-VL outperforms other models in terms of both accuracy and \dcheck{interpretability}. The prompt is designed to guide the model towards concise and reasoning-oriented answering. The template is as follows.

\begin{quote}
\textit{You are provided with a creative advertisement video. Please watch the video carefully and answer the given question based on its visual content and inferential reasoning. Limit your response for each question to no more than 30 words.} \\
\textit{Question: {}}
\end{quote}

This prompt encourages the model to perform visual understanding and inference while maintaining a concise response format. The video frames are extracted at a rate of 1 FPS. Processing long videos at 1 FPS will exceed the token limit of the GPT API. 
Thus, for GPT-4o, 30 frames per video are instead uniformly sampled.

Three observations are made: (1) The Qwen2.5-VL model (score=0.55 on the 72B model, 0.51 on 32B, 0.46 on 7B) outperforms other models, including GPT-4o (score=0.50), demonstrating superior capability in reasoning tasks. 
(2) The expansion of parameter count in the Qwen2.5-VL model family demonstrates a positive correlation with improvements in both video understanding and reasoning capabilities.
(3) The audio information in Qwen2.5-Omni fails to yield measurable improvements (score=0.44 on 7B). This is likely because the audio in advertisement videos consists primarily of background music, which offers limited semantic cues for reasoning.

\subsubsection{T-STAR}

(submitted by \textit{ActiveAlphaAgent} Team)

This solution T-STAR (\underline{t}est-time \underline{s}trategy \underline{t}uning and \underline{a}daptive \underline{r}easoning)
focuses on optimizing the inference strategy of the base model Qwen2.5-VL-72B-Instruct.

\paragraph{Inference strategy.}
This efficient approach directly enhances the model's prompt perception and response through three key hyper-parameters.

\noindent \textbf{(1) Video sampling rate.} 
Multiple frame rates (2, 3, and 5 FPS) are systematically evaluated to identify the optimal balance between motion detail and processing efficiency.

\noindent \textbf{(2) Frame resolution.} 
Various resolutions are assessed to achieve a balance between providing adequate visual detail for the visual encoder and maintaining efficiency.

\noindent \textbf{(3) Rejection sampling.} 
 Any response exceeding 30 words is discarded and regenerated. To enforce this, a strict instruction is prepended to the prompt as
\textit{``REMEMBER: Your answer MUST be 30 words or fewer. Question: \{\textit{original\_question}\}''.}

The team's top-performing leaderboard submission uses a configuration of FPS=2 while maximizing frame resolution. A key insight is that for the abstract reasoning in VR-Ads, prioritizing high spatial resolution, even at the cost of a lower frame rate, proves more effectively than increasing temporal sampling. This approach yields better performance under inference constraints, suggesting that critical visual cues in ads depend more on spatial detail than on high-frequency temporal information.

\subsubsection{QAAF}

\begin{figure*}[!h]
\centering
\includegraphics[width=\textwidth]{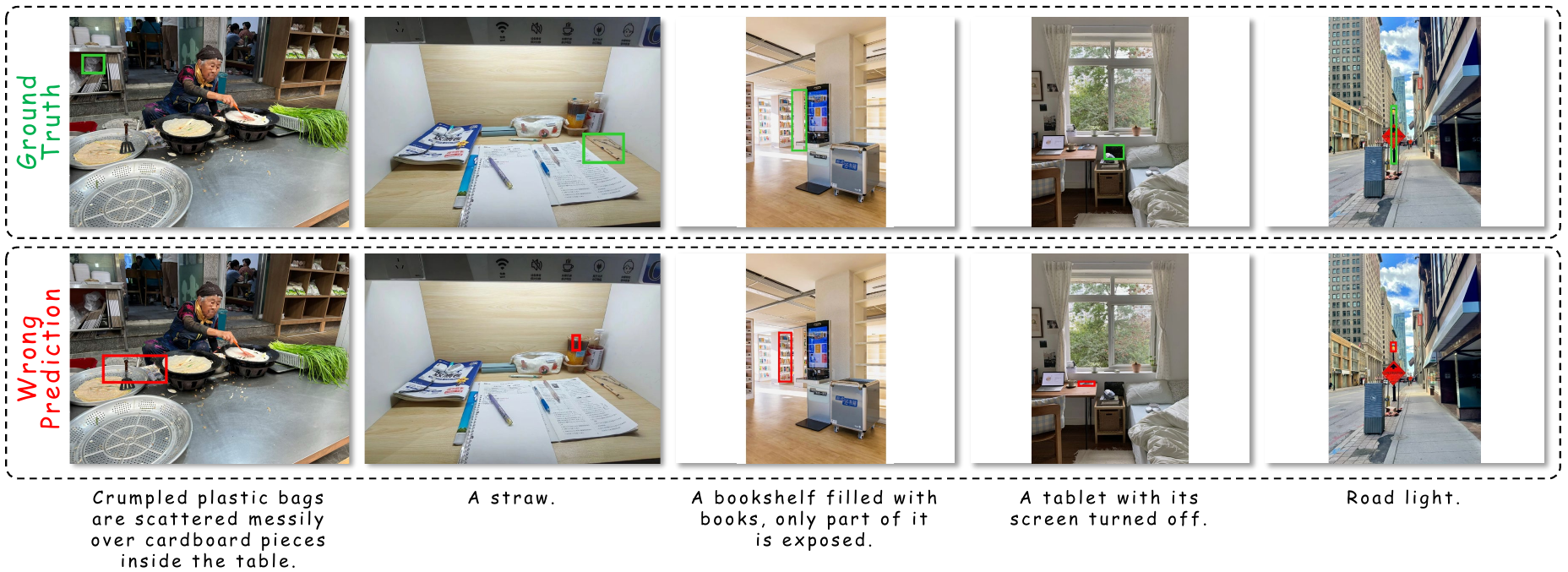} 
\caption{Failure cases of Track \#1 winner on \lens. 
Images are shown in their original proportions.
}
\label{fig_t1_fail_samples}
\end{figure*}

(submitted by \textit{rookiesllm} Team)

The team presents a \underline{Q}wen-based \underline{a}dvertising \underline{a}nalysis \underline{f}ramework (QAAF). 
To perceive rich information in advertising videos~\cite{Guo_2021}, the framework incorporates a dedicated preprocessing pipeline where an audio extraction module isolates and extracts spoken audio content from raw videos. 
The extracted speech is translated into English to standardize the language for subsequent analysis, ensuring robustness across multilingual advertisements. 
The Qwen2.5-VL-72B-Instruct model processes the multimodal input, which combines original video frames with standardized English text derived from the audio, integrating these synchronized visual and auditory cues to produce detailed, accurate, and contextually relevant responses to complex queries.

\paragraph{Implementation details.} 
Key parameters are set to a video sampling rate of 2 FPS and a maximum generation length of 2,048 new tokens. 
The results of automatic speech recognition are integrated via structured prompts. 
The voiceover content is explicitly prefixed with \textit{``Voiceover:''} to disambiguate it from the visual information, thus enabling the correlation between auditory and visual features.
A tailored prompt template drives the model to advertising-specific analysis.
\begin{quote}
\textit{You are an expert in video analysis and advertising content understanding. Analyze the provided video frames and answer accurately based on visual content, focusing on key elements, actions, and messages conveyed.}
\end{quote}

\subsubsection{GLM4ads}

(submitted by \textit{mm618} Team)

The team finetunes the GLM-4.1V-Thinking-9B model~\citep{hong2025glm} on the Video-Holmes dataset~\citep{cheng2025video}
that includes 270 manually annotated suspense short films and 1,837 questions covering seven reasoning types. Each question is accompanied by detailed answer explanations.
To enhance the model's reasoning capability, the team creates ``think-then-answer'' training samples by designing structured reasoning prompts for question-answer pairs.
\begin{quote}
\textit{You are an expert video analyst. Please watch the provided video and generate a coherent reasoning process based on the given correct answer and explanation. Your thought process should explain how the answer was derived, incorporating analysis of the video content.}
\end{quote}

\section{Discussion and Outlook}
\label{sec:discussion_outlook}

\begin{figure*}[!h]
\centering
\includegraphics[width=\textwidth]{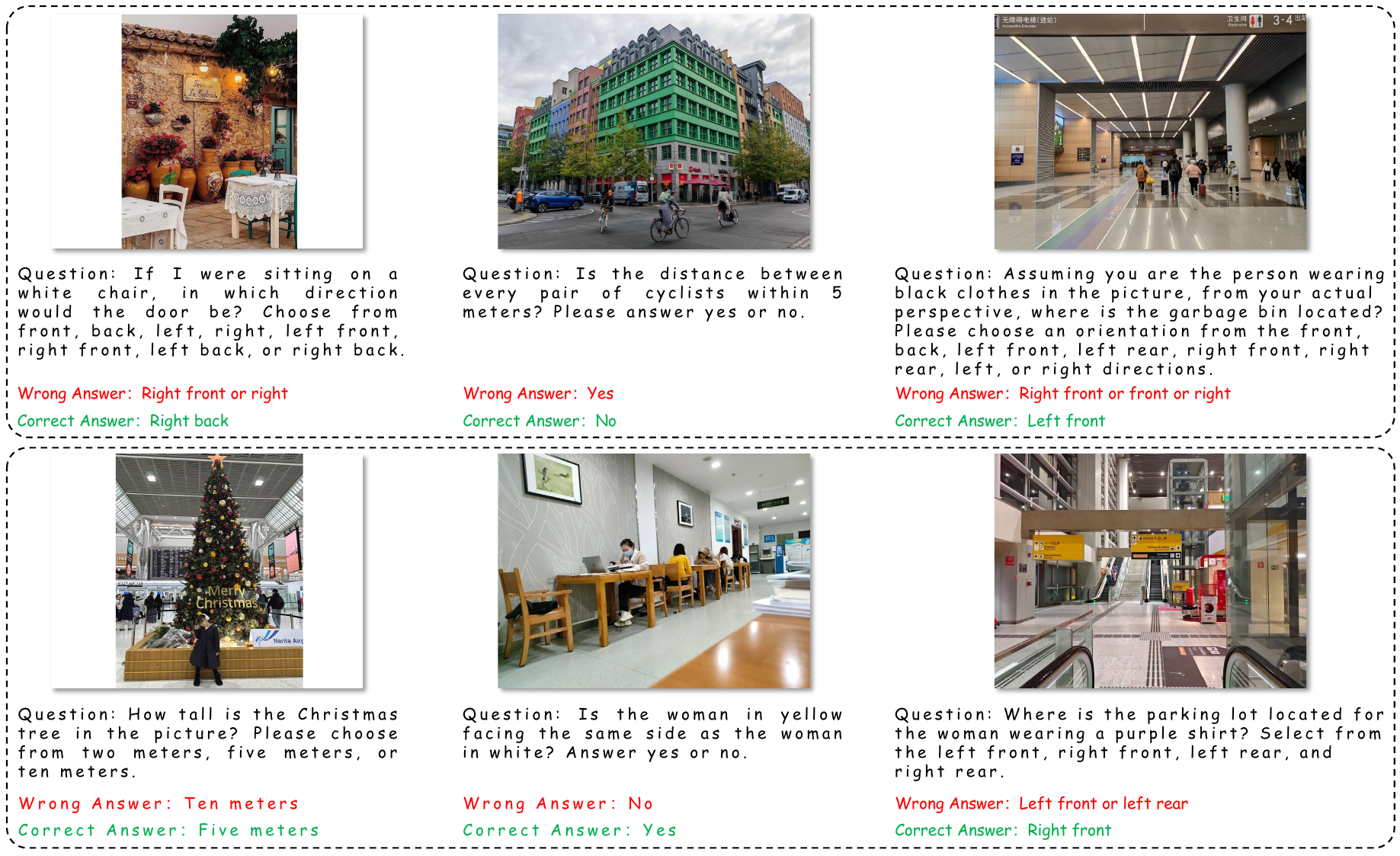} 
\vspace{-0.7cm}
\caption{Failure cases of Track \#2 winner on \lens. Images are shown in their original proportions.}
\label{fig_t2_fail_samples}
\end{figure*}

 \begin{figure*}[!h]
    \centering
    \includegraphics[width=\textwidth]{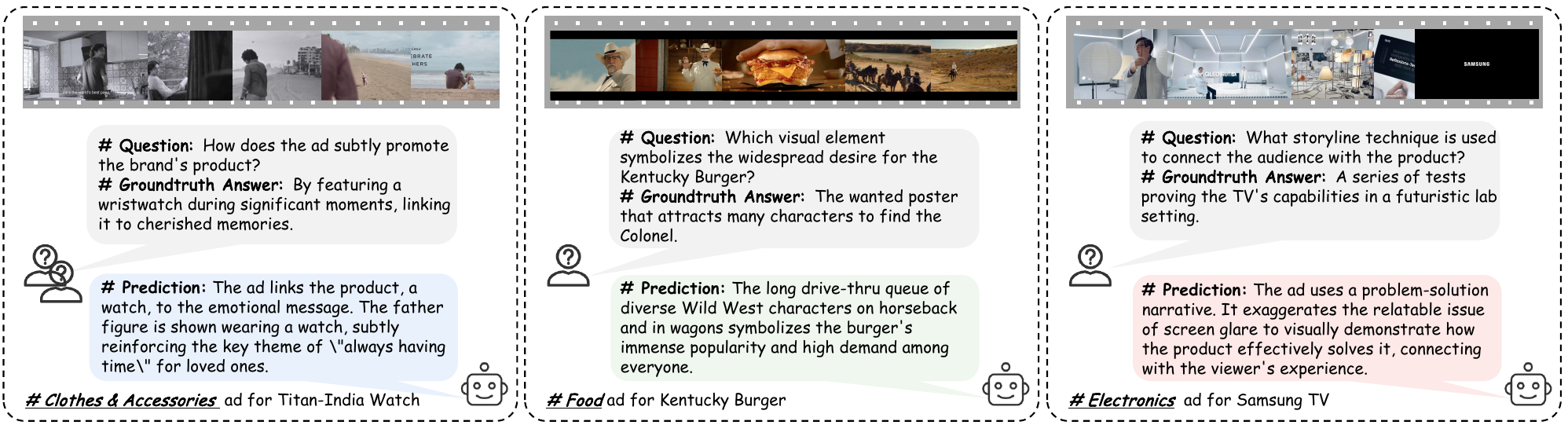}
    \vspace{-0.7cm}
    \caption{Failure cases of Gemini 2.5 Pro on \adsqa{}. 
    }
    \label{fig:track3-hard-examples}
\end{figure*}

 \begin{figure*}[!h]
    \centering
    \includegraphics[width=\textwidth]{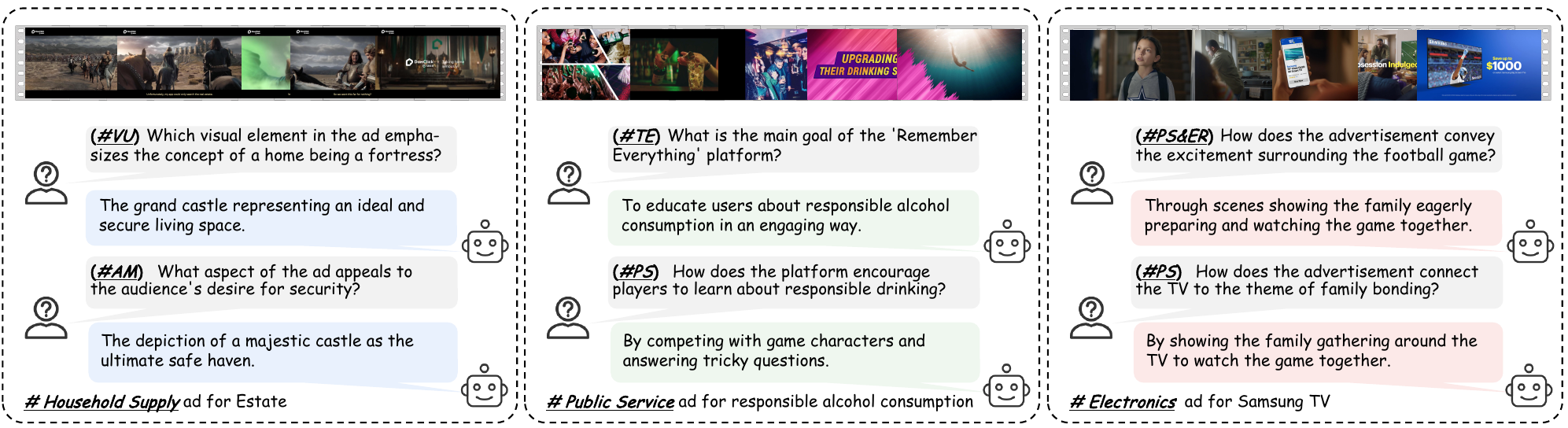}
    \vspace{-0.7cm}
    \caption{Samples of Track \#3, demonstrating the key challenges in advertisement understanding, \eg, visual elements, persuasive strategies, core themes, emotional appeals, and target audience analysis.
    }
    \label{fig:track3-examples}
\end{figure*}

\subsection{Further Discussion}

The committee has collected questions and feedback from the participants and the community. Based on the above results and methods, we would like to discuss the following open problems.

\noindent \textbf{(1) What are the strengths and weaknesses of the submitted methods?}

The submitted solutions have achieved performance gains based on strong LLM base models, integrating various LLM techniques, \eg, data augmentation, prompt engineering, SFT, and RL training.
However, one of the main goals of developing artificial intelligence methods is to create trustworthy, responsible, and generalizable AI. These methods would seem to still lack reliability and generalization capability.
For instance, as discussed in Section~\ref{sec:methods_results}, teams such as \textit{ActiveAlphaAgent}, \textit{Location depends on guessing}, and \textit{CV and RL}, use RL to conduct  task-oriented alignment training for LLMs, only considering IoU scores as their reward function. 
This reward signal is suboptimal as it only considers the task scores and likely drives the models to lose other capabilities.

\noindent \textbf{(2) Do LLMs already work well for multimodal reasoning?}

Multimodal reasoning in real-world scenarios and specialized domains is a challenging task.
Figure~\ref{fig_t1_fail_samples} and Figure~\ref{fig_t2_fail_samples} show randomly selected failure cases of the winning methods of Track \#1 and Track \#2, respectively. 
Two primary limitations are observed. First, in fine-grained image understanding, current multimodal large language models often confuse similar materials (\eg, mistaking plastic bags for other objects) or struggle to identify semantic targets (\eg, distinguishing a bookshelf filled with books). Second, in visual question answering, models exhibit notable biases in perspective understanding.
These failure cases suggest that the models tend to overly rely on first-person perspective priors and misinterpret concepts such as ``distance'', reflecting a lack of physical commonsense and highlighting the difficulty in reasoning about real-world scenes.

For Track \#3, its domain-specific challenges mainly include non-physical content reasoning, \eg, marketing logic, persuasive strategies, and audience engagement.  To highlight how challenging this kind of implicit content based reasoning is, we use our \adsqa{} dataset to evaluate the state-of-the-art MLLM Gemini 2.5 Pro.
Figure~\ref{fig:track3-hard-examples} presents three randomly selected failure cases.
It is observed that this state-of-the-art MLLM is nevertheless prone to generating suboptimal answers as well as content hallucination.
The model omits some details of the advertisements and incorporates some inherent biases.

\subsection{Outlook}

Developing general artificial intelligence is of great significance, and at the same time, studying LLMs tailored for specific scenarios and specialized fields is also crucial.
Both generalist and specialist models are useful.
Data and evaluation are crucial elements in driving LLMs and multimodal learning forward. 
Our vision is to explore more real-world scenarios and specialized domains, while also collecting more high-quality data as the testbeds for multimodal reasoning research.
In particular, as shown in Figure~\ref{fig:track3-examples}, we will further explore the fine-grained specialized logic and details in advertisement videos, including but not limited to persuasive strategies, core themes, and emotional appeals.
For the next challenge, we will provide more tracks and upgrade the scale of our test sets.

\section{Conclusion}
\label{sec:conclusion}

This report has reviewed the MARS2 2025 Challenge on Multimodal Reasoning.
Our goal is to bring together perspectives from multiple disciplines and to help researchers follow the state-of-the-art in both multimodal machine learning and multimodal large language models.
Given the remarkable progress of LLMs and multimodal machine learning, it is time to explore more challenging and specialized tasks that require complex multimodal reasoning and slow thinking towards ``System 2''.

MARS2 2025 challenge focuses on real-world scenarios to broaden the multimodal reasoning applications of MLLMs.
In particular, we explore two challenges, \ie, synergistic effects among reasoning tasks and non-stepwise complex reasoning.
Thus, we released two tailored datasets \lens{} and \adsqa{} as our competition test sets that support \dcheck{general reasoning} in 12 daily scenarios and \dcheck{domain-specific reasoning} in advertisement videos, respectively.
Three competition tracks are provided, \ie, Visual Grounding in Real-world Scenarios (VG-RS), Visual Question Answering with Spatial Awareness (VQA-SA), and Visual Reasoning in Creative Advertisement Videos (VR-Ads).

This year's challenge lasted for more than two months. 
\dcheck{76} teams from renowned academic and industrial institutions (\eg, \dcheck{ByteDance, NVIDIA, and Samsung}) have registered and submitted \dcheck{1200+} entries in total, of which \dcheck{40+} submissions have been included in our final ranking lists.

In addition, our committee has evaluated \dcheck{40+} baselines, including generalist MLLMs and task-specific models. 
The sizes of the baselines and the participants' models range from \dcheck{3B to 72B}. 
Thus, the scales of our datasets and results form a large and comprehensive benchmark for the field of multimodal reasoning.
\dcheck{This benchmark provides a comprehensive comparison covering ``generalist models \vs specialist models'', ``large models \vs small models'', ``open-source models \vs closed-source models'', \etc.}

The submitted solutions involve open-source models and commercial LLMs, and present some shared ideas that mainly include using ensembles, data augmentation, prompt engineering, alignment training, \etc.
To tackle complex multimodal reasoning, most teams adopted state-of-the-art MLLMs as their base models and then conducted multi-step alignment by using SFT and RL.
The collaboration of generalist and specialist models 
is widely used and achieves good performance in our challenges.

Our datasets (\lens{} and \adsqa{}), code sets (\dcheck{40+} baselines and \dcheck{15+} participants' methods), and rankings are publicly available on the MARS2 workshop website and our GitHub organization page \url{https://github.com/mars2workshop}, where our updates and announcements of upcoming events will be continuously published. 
Future work will continue to focus on multimodal reasoning, and we are committed to providing new application scenarios and high-quality data. We will continue to strive towards the goal of providing a more comprehensive benchmark and promoting the open-source community in the field of multimodal reasoning.

\section*{Acknowledgements}
The organizing team is \dcheck{in part} supported by the National Key Research and Development Program of China \dcheck{No.2022ZD0160600}.
Peng Xu is also supported by NSFC No.62306162.
We thank the MARS2 2025 sponsors Interdisciplinary AI Research Institute of Wuhan College, and iFLYTEK.
All the submissions were evaluated by an open source platform EvalAI.

{
    \small
    \bibliographystyle{ieeenat_fullname}
    \bibliography{main-simplified}
}

\clearpage

\appendix

\section{Related Work}
\label{sec:relat}
\subsection{Multimodal Reasoning with Large Vision-Language Models.}
Recent advances in LLMs have markedly improved their reasoning performance in text-only tasks. 
Nevertheless, owing to modality heterogeneity and the inherent complexity of cross-modal reasoning, extending these capabilities to MLLMs is still in an early exploratory phase. 
Current research predominantly follows three paradigms: multimodal chain-of-thought, reinforcement learning based reasoning, and native multimodal architectures.

\noindent \textbf{Multimodal chain-of-thought (M-CoT).}
Representative M-CoT approaches such as Multimodal-CoT \cite{zhang2023multimodal}, LLaVA-CoT \cite{xu2024llava}, RAVEN-CoT \cite{zhu2025data}, Magma-CoT \cite{yang2025magma}, and MM-ReAct \cite{yang2023mm} improve performance by explicitly constructing reasoning steps within visual prompts. 
They achieve notable accuracy gains on datasets like ScienceQA \cite{saikh2022scienceqa} and OK-VQA \cite{marino2019ok}, and show clear control over reasoning trajectories in tasks such as VG-RS and VQA-SA. 
However, these methods rely heavily on explicit visual cues and often falter when handling metaphors, emotional undertones, or abstract semantics common in advertising videos, where reasoning chains can become vague or ambiguous.

\noindent \textbf{Reinforcement learning based reasoning.}
The RL-based reasoning optimization paradigm, exemplified by models such as Visual-RFT \cite{liu2025visual}, SRPO \cite{wan2025srpo}, and VL-Rethinker \cite{wang2025vl}, explicitly refines the reasoning trajectories of multimodal models through reward functions grounded in visual and textual cues, group-wise relative policy optimization, and mechanisms of self-reflection and soft replay. 
This paradigm has demonstrated strong performance in tasks like VQA-SA, which require spatial relation understanding and multi-step logical reasoning. 
Nevertheless, its reliance on explicit reward signals and high-quality feedback poses challenges for establishing effective supervision in abstract reasoning scenarios such as VR-Ads of our Track \#3, thus constraining its capacity to generalize.

\noindent \textbf{Native multimodal architectures.}
Native multimodal reasoning architectures enable effective fusion and reasoning across modalities through unified model designs and multimodal pretraining mechanisms. 
Representative closed-source models, including Gemini \cite{team2024gemini}, GPT-4V \cite{yang2023dawn}, and Claude \cite{anthropic2024model}, exhibit inherent end-to-end multimodal reasoning capabilities and demonstrate robust performance across diverse task scenarios. 
In parallel, open-source models such as LLaVA \cite{liu2023visual} and MiniGPT-4 \cite{zhu2023minigpt} leverage lightweight architectures to achieve broad multimodal reasoning capacity, showing competitive performance in visual grounding and spatial reasoning. Nevertheless, these architectures still encounter significant challenges in abstract conceptual reasoning tasks, such as VR-Ads.

\noindent \textbf{Summary.} Although recent MLLMs such as GPT-4o \cite{achiam2023gpt}, Gemini 2.5 Pro \cite{team2024gemini}, InternVL3 \cite{zhu2025internvl3}, and Qwen2.5-VL \cite{bai2025qwen2} have demonstrated strong performance in multimodal reasoning, the \texttt{Lens} benchmark reveals that they continue to underperform in spatial reasoning, compositional understanding, and abstract reasoning, particularly in hierarchical reasoning tasks. 
This highlights the limitations of current models in fundamental visual perception and logical reasoning, as well as the shortcomings of fragmented evaluation systems that fail to provide a systematic characterization of the continuum from perception to reasoning.
To address this gap, the MARS2 competition introduced a unified evaluation framework encompassing three complementary tracks: VG in Physical Scenes (VG-RS), VQA with Spatial Awareness (VQA-SA), and VR in Creative Advertisement Videos (VR-Ads). 
This initiative offers a more comprehensive benchmark platform to advance the reasoning capabilities of next-generation MLLMs.

\subsection{Visual Grounding and Scene Perception}
Visual Grounding (VG) refers to the task of identifying regions of interest within an image based on textual descriptions. 
By analyzing image scenes, VG establishes cross-modal associations between vision and language, thereby enabling machines to achieve human-like multimodal understanding and reasoning. 
With the rapid development of LLMs and MLLMs, research on VG has evolved from traditional task-specific models towards perception models that leverage the inherent reasoning capacities of large models. 
Currently, LLM-based VG can be broadly categorized into three research paradigms: integration of external visual information, cross-modal semantic alignment, and multimodal reasoning strategies.

\noindent \textbf{Integration of external visual information.}
VG takes both text and image modalities as input, and LLMs alone are inadequate for accomplishing this task. 
To address this limitation, several studies have incorporated external visual encoders to process image information. 
Some approaches emphasize region-level visual representations, for instance, Groma \cite{ma2024groma} employs DINO \cite{zhang2022dino} as the image encoder to capture key regions and applies explicit modeling with region tokens to leverage the spatial reasoning capabilities of MLLMs for localization. 
To support scene perception tasks at varying levels of granularity, the external visual encoders in Ferret \cite{you2024ferret} and Ferret-v2 \cite{zhang2024ferret} employ a multi-scale visual backbone to enhance spatial semantics via multi-scale visual sampling. 
Furthermore, methods such as Groundhog \cite{zhang2024groundhog} and GLaMM \cite{rasheed2024glamm} enable mask generation as an alternative to bounding boxes for localization.

\noindent \textbf{Cross-modal semantic alignment.}
Most external visual encoder approaches rely on static perception mechanisms, which limit their ability to disentangle fine-grained semantics guided by complex textual inputs. 
Cross-modal semantic alignment methods instead establish a unified semantic space through learnable queries, adapter modules, or contrastive learning, offering greater adaptability to linguistic variation. 
A representative paradigm is Q-Former-based fusion, such as BLIP-2 \cite{li2023blip} and Grounding-GPT \cite{li2024groundinggpt}. They employ Q-Former to bridge vision and language, constructing query-guided visual representations that enhance semantic consistency across modalities and improve localization alignment capabilities. 
To integrate multi-scale visual information, LION \cite{chen2024lion} introduces a Mixture-of-Adapters framework, using multi-granularity adapters combined with a router mechanism for fusion within the LLM. 
As grounding requirements become more sophisticated, approaches such as Llava-grounding \cite{zhang2024llava} and Grounding-GPT \cite{li2024groundinggpt} enable mixed inputs and outputs in the form of bounding boxes, textual annotations, and masks, thereby laying the foundation for unified interfaces that support interactive multimodal dialogue.

\noindent \textbf{Multimodal reasoning strategies.}
With reasoning strategies in LLMs achieving significant breakthroughs across various tasks, multimodal reasoning strategies have also been applied to VG. 
Shikra \cite{chen2023shikra} extends the traditional CoT framework by introducing Grounding CoT, which incorporates region referencing and localization factors to improve semantic reference resolution. 
To further strengthen a model’s ability to follow linguistic instructions, multi-stage instruction tuning has been adopted in VG. 
Representative approaches, such as Kosmos-2 \cite{peng2023kosmos}, employ a three-stage paradigm of ``perception–instruction tuning–alignment" to progressively improve cross-modal semantic alignment.
Building on this foundation, recent studies have further extended multi-instruction tuning strategies. 
VPP-LLaVA \cite{tang2025visual} constructed VPP-SFT, a high-quality grounding instruction dataset comprising approximately 600k samples, and applied it for multi-instruction tuning on the LLaVA family of models, substantially improving controllability in coordinate-to-region outputs. 
TRIG \cite{li2025towards} expands VG from natural images to text-rich images such as forms, documents, and receipts, proposing novel tasks along with a specialized instruction dataset. 
In the Doc-VQA setting, it achieves localized answer alignment, thereby filling a critical gap of traditional VG in document understanding. 
VideoGLaMM \cite{munasinghe2025videoglamm} focuses on video-based multi-instruction tuning, significantly enhancing spatiotemporal alignment and pixel-level localization, and extending multi-instruction paradigms to video tasks. 
LLaVA-ST \cite{li2025llava} unifies temporal localization, spatial localization, and descriptive tasks under a single framework for multi-task and multi-instruction training.
Meanwhile, VistaLLM \cite{pramanick2024jack} strengthens cross-modal transfer and enhances generalization by engaging in joint grounding tasks across multiple modalities, including images, videos, and audio.
LLaVA-c \cite{liu2025llava} is designed for continuously expanding multi-task instruction training scenarios, aiming to mitigate the problem of catastrophic forgetting induced by task expansion.

\noindent \textbf{Summary.}
Current VG tasks primarily depend on established benchmark datasets such as RefCOCO \cite{chen2025revisiting} / RefCOCO+ \cite{chen2025revisiting} / RefCOCOg \cite{mao2016generation}, ReferItGame \cite{kazemzadeh2014referitgame}, and RefCLEF \cite{kazemzadeh2014referitgame}, while some studies have also been evaluated on open-domain benchmarks like Flickr30K Entities \cite{plummer2015flickr30k} and Visual Genome \cite{krishna2017visual}. 
However, these datasets are insufficient to meet the evaluation requirements of MLLM reasoning. 
To bridge this gap, researchers have proposed more sophisticated benchmarks, including GQA \cite{hudson2019gqa} for scene-graph-based question answering, VCR \cite{zellers2019recognition} for region grounding and multi-turn QA, and NLVR$^2$ \cite{suhr2017corpus} for compositional alignment of visual concepts. 
In addition, benchmarks such as LLaVA-Bench \cite{liu2023visual}, MME \cite{yuan2025mme}, and MMBench \cite{liu2024mmbench} are designed to evaluate the zero-shot and few-shot generalization abilities of MLLMs, as well as their capacity to leverage external knowledge~\cite{long2024trust,long2024generative}.

Although existing benchmarks provide validation platforms for multimodal reasoning, most fail to evaluate spatial relationship modeling capabilities. 
To address this gap, the \texttt{Lens} Benchmark is constructed with a focus on spatial reasoning, designing multi-turn dialogue scenarios in which models must not only identify targets but also perform reasoning by integrating contextual and cross-modal information. 
Unlike prior benchmarks that primarily relied on MS COCO or Flickr, \texttt{Lens} draws on real-world data with human annotations, over 80\% of which were collected after September 2024, ensuring both temporal relevance and robust generalization to unseen data. 
Experimental results demonstrate that even advanced MLLMs such as GPT-4o \cite{achiam2023gpt}, Gemini 2.5 Pro \cite{team2024gemini}, InternVL3 \cite{zhu2025internvl3}, and Qwen2.5-VL \cite{bai2025qwen2} continue to encounter significant challenges in spatial reasoning tasks. 
Consequently, \texttt{Lens} constitutes an important complement for assessing the depth of multimodal reasoning chains and spatial perception capabilities.

\subsection{Visual Question Answering with Spatial and Commonsense Reasoning}
Visual Question Answering (VQA) serves as a cornerstone for achieving human-like multimodal reasoning, while spatial perception underpins human understanding of the physical world. 
Consequently, VQA research has increasingly focused on spatial perception, seeking to capture the physical structures and spatial relationships embedded in images. 
With the advent of vision–language pre-trained models such as ViLBERT \cite{lu2019vilbert}, VisualBERT \cite{li2019visualbert}, and VL-BERT \cite{su2019vl}, VQA has made remarkable advances. 
These models acquire joint visual–language representations via large-scale pretraining on image–text pairs, followed by fine-tuning on tasks such as VQA, leading to substantial gains in accuracy. 
In this section, we review four representative paradigms: explicit spatial relationship modeling, implicit spatial-aware feature fusion, and spatial alignment and CoT reasoning enhancement.

\noindent \textbf{Explicit spatial relationship modeling.}
Explicit approaches construct structured representations of object relations based on geometric priors such as relative position, orientation, and distance, and integrate them into LLM reasoning chains to enable symbolic or structured reasoning. 
XNM \cite{shi2019explainable}, NS-VQA \cite{yi2018neural}, and Neural State Machine \cite{hudson2019learning} all generate object–relation graphs from bounding box coordinates and semantic categories, thereby supporting interpretable reasoning over spatial layouts. 
Scene Graph Reasoning for VQA \cite{hildebrandt2020scene} further incorporates both semantic and spatial edges to address complex queries, while SelfGraphVQA \cite{de2023selfgraphvqa} introduces self-supervised graph augmentation to reduce annotation costs. 
The GQA dataset \cite{hudson2019gqa}, which provides annotations including scene graphs and functional programs, has become the standard benchmark for this class of methods.

\noindent \textbf{Implicit spatial-aware feature fusion.}
In contrast to explicit modeling of spatial information, implicit fusion methods leverage cross-modal attention mechanisms and multi-scale feature integration, allowing models to learn spatial dependencies implicitly during training. 
Early cross-modal pretraining models, such as LXMERT \cite{tan2019lxmert}, UNITER \cite{chen2020uniter}, and OSCAR \cite{li2020oscar}, incorporate self-attention and cross-attention mechanisms to align textual tokens with localized visual representations, thereby implicitly capturing spatial patterns.
Subsequently, VinVL \cite{zhang2021vinvl} enhances the visual encoder with object-centric features, while MDETR \cite{kamath2021mdetr} adopts a text-modulated detection framework to jointly model text and images in an end-to-end reasoning process. 
More recently, QA-ViT \cite{ganz2024question} directly injects question embeddings into the visual encoder, directing multi-scale attention towards regions pertinent to the question and thereby capturing both local and global spatial relationships effectively.

\noindent \textbf{Spatial alignment and CoT reasoning enhancement.}
These methods strengthen vision–language spatial alignment and enable multi-step reasoning. 
Region-level alignment, such as CLOC \cite{chen2410contrastive}, integrates region–text contrastive learning into CLIP, improving the mapping between spatial descriptors (\textit{e.g.}, ``top left corner'') and image regions, which is especially useful for queries with spatial prepositions. 
Meanwhile, CoT prompting decomposes complex spatial reasoning into interpretable steps, with resources like Visual-CoT \cite{shao2024visual} providing bounding box and reasoning annotations for supervised training.

\noindent \textbf{Summary.}
In VQA research with spatial perception capabilities, the design and selection of benchmark datasets directly affect the evaluation and comparison of model capabilities. 
Current mainstream benchmarks encompass both real-world and synthetic environments, as well as diverse forms across two-dimensional and three-dimensional spaces. 
This highlights the dual requirements of spatial reasoning tasks for dataset diversity and controllable difficulty.
In two-dimensional real-world image scenarios, GQA \cite{hudson2019gqa} provides large-scale annotations of attributes and relations, extensively covering relative positions (\textit{e.g.}, ``X is to the left of Y"), making it well-suited for evaluating multi-step spatial reasoning. 
The spatial subset of VQA v2 \cite{goyal2017making} is more closely aligned with general VQA tasks. Although relatively limited in scale, it enables testing of spatial generalization in open-domain settings. 
TDIUC \cite{kafle2017analysis} introduces a dedicated spatial-relation subtask, facilitating cross-task comparisons of model capabilities.
In synthetic environments, CLEVR \cite{johnson2017clevr} and CLEVR-Humans \cite{holzinger2023toward} are designed to test spatial reasoning via controllable attribute combinations and geometric relationships.
CLEVR-Humans \cite{holzinger2023toward} further incorporates human-authored natural language questions, increasing linguistic diversity and naturalness, and is often employed to analyze the effectiveness of specific modeling strategies.
In three-dimensional contexts, ScanQA \cite{azuma2022scanqa} and 3D-VQA \cite{9866910,9984686} extend VQA to indoor and outdoor 3D scans and point clouds, requiring models to reason over 3D structures and object layouts. 
These benchmarks are highly relevant to applications in robotic navigation and AR/VR systems, while also assessing the integration of depth information and the learning of 3D representations.

Compared with conventional VQA benchmarks, the \texttt{Lens} benchmark evaluates spatial reasoning ability from multiple dimensions by integrating tasks such as spatial relation understanding, visual reference resolution, and 3D scene comprehension. 
Its question design goes beyond object localization to also include spatial relation inference and multi-step reasoning, resulting in a more comprehensive spatial reasoning chain. 
This provides a more faithful assessment of a model’s capacity to capture complex spatial semantics than traditional benchmarks that focus solely on positional relations. 
Moreover, in contrast to synthetic datasets, \texttt{Lens} is more closely aligned with real-world application scenarios such as robotic navigation, AR/VR, and cross-view retrieval, thereby mitigating the domain gap between synthetic and real data.

\subsection{Visual Reasoning in Creative Advertisement Videos}

Creative advertising video reasoning seeks to exploit multimodal information within advertisements to uncover the implicit abstract semantics underlying explicit visual elements. 
In contrast to short-duration factual datasets such as MSVD-QA \cite{chowdhury2018hierarchical} or TGIF-QA \cite{jang2017tgif}, and long-duration narrative datasets such as TVQA \cite{lei2018tvqa} and MovieQA \cite{tapaswi2016movieqa}, creative advertising videos are shorter in duration yet contain richer abstract cues. 
These videos place a stronger emphasis on abstract reasoning, including emotion analysis, metaphor comprehension, and cultural symbol interpretation. 
Thus, the essence of the reasoning process lies not in recognizing explicit visual facts, but rather in integrating multimodal evidence to infer latent thematic meanings.

\noindent \textbf{Advertisement video understanding benchmarks.}
Early video question answering benchmarks, such as MSRVTT-QA \cite{xu2017video} and TGIF-QA \cite{jang2017tgif}, primarily emphasize action recognition and temporal sequencing. They are designed to evaluate fundamental perception and short-term reasoning abilities.
Narrative-oriented datasets like \cite{lei2018tvqa} and MovieQA \cite{tapaswi2016movieqa} extend the scope to multi-turn, dialogue-based reasoning requiring long-term temporal dependencies. 
More recent causal and compositional reasoning benchmarks, including NExT-QA \cite{xiao2021next}, AGQA \cite{grunde2021agqa}, and STAR \cite{wu2024star}, focus on assessing models' ability to answer ``why" and ``how" questions, connecting visual evidence with underlying intentions and outcomes. 
While these benchmarks provide valuable references for the advancement of reasoning strategies, they are still limited in assessing abstract semantics and creative content.

In the advertising domain, the understanding of advertisements has emerged as a critical research direction due to the economic value of accurate content interpretation. 
Internet companies generate substantial revenue by automatically delivering ads to target audiences, thereby highlighting the necessity of automated advertisement understanding \cite{jia2023kafa}. 
The earliest Pit dataset \cite{hussain2017automatic} formalizes this task as a VQA problem \cite{long2025retrieval}, and is subsequently extended or adapted in later studies \cite{kalarani2024seeing} or private datasets \cite{yang2024synchronized}. These studies typically focus on specific dimensions, such as persuasion strategy analysis in image advertisements \cite{kumar2023persuasion, ye2019interpreting}, image ad retrieval \cite{zhao2024enhancing}, intent understanding \cite{jia2021intentonomy}, and visual metaphor comprehension \cite{akula2023metaclue, xu2024exploring, zhang2021multimet}. 
Although Pit primarily targets image advertisements, it also includes a subset of video ads. 
Nonetheless, Pit faces challenges of limited accessibility, insufficient diversity, and a constrained QA format, and to date, there remains no widely adopted comprehensive video QA benchmark for ads.

\noindent \textbf{Summary.}
To address this gap, \texttt{AdsQA} has recently been introduced as a dedicated benchmark for creative advertisement understanding. 
It evaluates five key dimensions: visual concept understanding, emotion recognition, theme and message identification, persuasion strategy analysis, and target audience recognition, capturing both low-level perceptual cues and high-level abstract semantics. 
Unlike general benchmarks such as MVBench \cite{li2024mvbench} or Video-MME \cite{fu2025video}, \texttt{AdsQA} explicitly frames the integration of visual, textual, and occasionally auditory modalities as a means to infer underlying communicative intent, rather than merely matching explicit visual facts. 
Through the VR-Ads track, it facilitates the transition of multimodal reasoning techniques from explicit perception tasks to higher-level cognitive reasoning. 
This raises greater demands on cross-modal fusion, temporal modeling, external knowledge integration, and interpretability, thereby providing a more comprehensive reflection of MLLMs’ capabilities in real-world multimodal reasoning scenarios. 
Moreover, it offers valuable implications for practical applications, including advertisement understanding, content moderation, user profiling, and recommendation systems.

\section{Teams and Affiliations}

\subsection{MARS2 2025 Challenge Organizing Team}
\noindent\textit{\textbf{Organizers: }} \\
Peng Xu$^1$ (\mhref{mailto:peng\_xu@tsinghua.edu.cn}{peng\_xu@tsinghua.edu.cn}) \\ 
Shengwu Xiong$^2$ (\mhref{mailto:xiongsw@whut.edu.cn}{xiongsw@whut.edu.cn}) \\
Jiajun Zhang$^3$ (\mhref{mailto:jjzhang@nlpr.ia.ac.cn}{jjzhang@nlpr.ia.ac.cn}) \\
Yaxiong Chen$^2$ (\mhref{mailto:chenyaxiong@whut.edu.cn}{chenyaxiong@whut.edu.cn}) \\ \\
\noindent\textit{\textbf{Steering Committee: }} \\
Bowen Zhou$^{1,4}$ (\mhref{mailto:zhoubowen@tsinghua.edu.cn}{zhoubowen@tsinghua.edu.cn}) \\
Chen Change Loy$^5$ (\mhref{mailto:ccloy@ntu.edu.sg}{ccloy@ntu.edu.sg}) \\
David A. Clifton$^6$ (\mhref{mailto:david.clifton@eng.ox.ac.uk}{david.clifton@eng.ox.ac.uk}) \\
Kyoung Mu Lee$^7$ (\mhref{mailto:kyoungmu@snu.ac.kr}{kyoungmu@snu.ac.kr}) \\
Luc Van Gool$^8$ (\mhref{mailto:luc.vangool@insait.ai}{luc.vangool@insait.ai}) \\ \\
\noindent\textit{\textbf{Contributors: }} \\
Ruiming He$^2$ (\mhref{mailto:361284@whut.edu.cn}{361284@whut.edu.cn}) \\
Ruilin Yao$^{2,3}$ (\mhref{mailto:yaoruilin@whut.edu.cn}{yaoruilin@whut.edu.cn}) \\
Xinwei Long$^1$ (\mhref{mailto:longxw22@mails.tsinghua.edu.cn}{longxw22@mails.tsinghua.edu.cn}) \\
Jirui Huang$^{2,3}$ (\mhref{mailto:284387@whut.edu.cn}{284387@whut.edu.cn})  \\
Kai Tian$^1$ (\mhref{mailto:tk23@mails.tsinghua.edu.cn}{tk23@mails.tsinghua.edu.cn}) \\
Sa Yang$^9$ (\mhref{mailto:yangsa2023@stu.pku.edu.cn}{yangsa2023@stu.pku.edu.cn})\\
Yihua Shao$^3$ (\mhref{mailto:yihuajerry@gmail.com}{yihuajerry@gmail.com})\\
Jin Feng$^{10}$ (\mhref{mailto:fengjin@kuaishou.com}{fengjin@kuaishou.com}) \\
Yue Zhong$^{11}$ (\mhref{mailto:zhongyue@cupl.edu.cn}{zhongyue@cupl.edu.cn})\\
Jiakai Zhou$^1$ (\mhref{mailto:zhoujk22@mails.tsinghua.edu.cn}{zhoujk22@mails.tsinghua.edu.cn}) \\
Cheng Tang$^1$ (\mhref{mailto:c-tang22@mails.tsinghua.edu.cn}{c-tang22@mails.tsinghua.edu.cn}) \\
Tianyu Zou$^2$ (\mhref{mailto:zoutianyu@whut.edu.cn}{zoutianyu@whut.edu.cn})\\
Yifang Zhang$^2$ (\mhref{mailto:332972@whut.edu.cn}{332972@whut.edu.cn})\\
Junming Liang$^2$ (\mhref{mailto:}{songlier@whut.edu.cn})\\
Guoyou Li$^2$ (\mhref{mailto:315863@whut.edu.cn}{315863@whut.edu.cn})\\
Zhaoxiang Wang$^2$ (\mhref{mailto:wangzx@whut.edu.cn}{wangzx@whut.edu.cn})\\
Qiang Zhou$^{12}$ (\mhref{mailto:amos@52cv.net}{amos@52cv.net})  \\
Yichen Zhao$^2$ (\mhref{mailto:zhaoyichen@whut.edu.cn}{zhaoyichen@whut.edu.cn})  \\
Shili Xiong$^2$ (\mhref{mailto:slxiong.illinois@gmail.com}{slxiong.illinois@gmail.com})\\
Hyeongjin Nam$^7$ (\mhref{mailto:namhj28@gmail.com}{namhj28@gmail.com}) \\
Jaerin Lee$^7$  (\mhref{mailto:ironjr@snu.ac.kr}{ironjr@snu.ac.kr})\\
Jaeyoung Chung$^7$  (\mhref{mailto:robot0321@snu.ac.kr}{robot0321@snu.ac.kr})\\
JoonKyu Park$^7$  (\mhref{mailto:jkpark0825@snu.ac.kr}{jkpark0825@snu.ac.kr})\\
Junghun Oh$^7$  (\mhref{mailto:dh6dh@snu.ac.kr}{dh6dh@snu.ac.kr})\\
Kanggeon Lee$^7$  (\mhref{mailto:dlrkdrjs97@naver.com}{dlrkdrjs97@naver.com})\\ 
Wooseok Lee$^7$  (\mhref{mailto:adntjr4@gmail.com}{adntjr4@gmail.com})\\
Juneyoung Ro$^{13}$ (\mhref{mailto:juneyoung.ro@kaist.ac.kr}{juneyoung.ro@kaist.ac.kr})\\ 
Turghun Osman$^{14}$ (\mhref{mailto:turghun@ms.xjb.ac.cn}{turghun@ms.xjb.ac.cn})\\ \\
\noindent\textit{\textbf{Affiliations: }} \\ 
$^1$ Tsinghua University \\
$^2$ Wuhan University of Technology \\
$^3$ Institute of Automation, Chinese Academy of Sciences \\
$^4$ Shanghai Artificial Intelligence Laboratory \\
$^5$ Nanyang Technological University \\
$^6$ University of Oxford \\
$^7$ Seoul National University \\
$^8$ INSAIT, Sofia University St. Kliment Ohridski \\
$^9$ Peking University \\
$^{10}$ KuaiShou Inc. \\
$^{11}$ ‌China University of Political Science and Law\\
$^{12}$ \dcheck{52CV Computer Vision Academic Community} \\
$^{13}$ Korea Advanced Institute of Science and Technology \\
$^{14}$ Xinjiang Technical Institute of Physics \& Chemistry of the Chinese Academy of Sciences\\

\subsection{MARS2 2025 Challenge Participants}
(alphabetically)
\subsubsection*{\textit{404 Not Found} Team}
\noindent\textit{\textbf{Members: }} \\
Jingxuan Li$^1$ (\mhref{mailto:liuy33467@gmail.com}{liuy33467@gmail.com})\\
Kai Tian$^2$ (\mhref{mailto:tk23@mails.tsinghua.edu.cn}{tk23@mails.tsinghua.edu.cn})\\
Sa Yang$^3$ (\mhref{mailto:yangsa2023@stu.pku.edu.cn}{yangsa2023@stu.pku.edu.cn}) \\
Jiaheng Ma$^4$ (\mhref{mailto:3220231772@bit.edu.cn}{3220231772@bit.edu.cn}) \\
Yang Liu$^5$ (\mhref{mailto:liuyang2015779381@gmail.com}{liuyang2015779381@gmail.com}) \\
\noindent\textit{\textbf{Affiliations: }} \\ 
$^1$ Harbin Engineering University \\
$^2$ Tsinghua University \\
$^3$ Peking University \\
$^4$ Beijing Institute of Technology \\
$^5$ China Electronics Technology Group Corporation \\
\subsubsection*{\textit{ActiveAlphaAgent} Team}
\noindent\textit{\textbf{Members: }} \\
Chong Peng$^1$ (\mhref{mailto:pengchong@meituan.com}{pengchong@meituan.com})\\
Taofeng Xue$^1$ (\mhref{mailto:xuetaofeng@meituan.com}{xuetaofeng@meituan.com})\\
Zijian Zhang$^1$ (\mhref{mailto:zhangzijian14@meituan.com}{zhangzijian14@meituan.com})\\
Jianing Wang$^1$ (\mhref{mailto:}{wangjianing16@meituan.com}) \\
Chengcheng Han$^1$ (\mhref{mailto:}{hanchengcheng02@meituan.com})\\
\noindent\textit{\textbf{Affiliations: }} \\ 
$^1$ Meituan-M17 \\
\subsubsection*{\textit{adaboost} Team}
\noindent\textit{\textbf{Members: }} \\
Ziang Li$^1$ (\mhref{mailto:24241215014@stu.xidian.edu.cn}{24241215014@stu.xidian.edu.cn})\\
Linnan Zhao$^1$ (\mhref{mailto:lnzhao@stu.xidian.edu.cn}{lnzhao@stu.xidian.edu.cn}) \\
Xinyi You$^1$ (\mhref{mailto:24171213968@stu.xidian.edu.cn}{24171213968@stu.xidian.edu.cn})\\
\noindent\textit{\textbf{Affiliations: }} \\ 
$^1$ Xidian University \\
\subsubsection*{\textit{CV and RL} Team}
\noindent\textit{\textbf{Members: }} \\
Feng Gao$^1$ (\mhref{mailto:gaofeng24@mails.ucas.ac.cn}{gaofeng24@mails.ucas.ac.cn})\\
Zongshu Li$^2$ (\mhref{mailto:lizongshu@wair.ac.cn}{lizongshu@wair.ac.cn}) \\
Hanxiao Wu$^{1,3}$ (\mhref{mailto:hxwu@whut.edu.cn}{hxwu@whut.edu.cn}) \\
Guibo Zhu$^{1,2}$ (\mhref{mailto:gbzhu@nlpr.ia.ac.cn}{gbzhu@nlpr.ia.ac.cn})\\
\noindent\textit{\textbf{Affiliations: }} \\ 
$^1$ Institute of Automation, Chinese Academy of Sciences \\
$^2$ Wuhan Artificial Intelligence Research \\
$^3$ Wuhan University of Technology \\
\subsubsection*{\textit{EAIC} Team}
\noindent\textit{\textbf{Members: }} \\
Nguyen Thanh Thien$^1$ (\mhref{mailto:thiennt@uit.edu.vn}{thiennt@uit.edu.vn})\\
\noindent\textit{\textbf{Affiliations: }} \\ 
$^1$ University of Information Technology, Ho Chi Minh City, Vietnam \\
\subsubsection*{\textit{Echoch} Team}
\noindent\textit{\textbf{Members: }} \\
Chenhao Qiu$^1$ (\mhref{mailto:qiuchenhao@mgtv.com}{qiuchenhao@mgtv.com})\\
Xusheng Liu$^1$ (\mhref{mailto:liuxusheng@mgtv.com}{liuxusheng@mgtv.com})\\
Can Hu$^1$ (\mhref{mailto:hucan@mgtv.com}{hucan@mgtv.com}) \\
Jie Yang$^1$ (\mhref{mailto:yangjie@mgtv.com}{yangjie@mgtv.com})\\
Shien Song$^1$ (\mhref{mailto:shien@mgtv.com}{shien@mgtv.com}) \\
Haibo Lu$^1$ (\mhref{mailto:haibo2@mgtv.com}{haibo2@mgtv.com}) \\
Han Qi$^1$ (\mhref{mailto:qihan@mgtv.com}{qihan@mgtv.com})\\
\noindent\textit{\textbf{Affiliations: }} \\ 
$^1$ MGTV Corporation, Changsha, China \\
\subsubsection*{\textit{gogogo\_truefaler} Team}
\noindent\textit{\textbf{Members: }} \\
Zhaohong Liu$^1$ (\mhref{mailto:liuzhaohong0425@cuc.edu.cn}{liuzhaohong0425@cuc.edu.cn})\\
Yiting Xi$^1$ (\mhref{mailto:17749772485@163.com}{17749772485@163.com}) \\
Zhenni Huang$^1$ (\mhref{mailto:hzn0806@163.com}{hzn0806@163.com}) \\
Ziyun Xiao$^1$ (\mhref{mailto:yolandaxiao@foxmail.com}{yolandaxiao@foxmail.com})\\
\noindent\textit{\textbf{Affiliations: }} \\ 
$^1$ Communication University of China \\
\subsubsection*{\textit{GoodAI\_zju} Team}
\noindent\textit{\textbf{Members:}} \\
Xin Chen$^1$ (\mhref{mailto:745482277@qq.com}{745482277@qq.com})\\
\noindent\textit{\textbf{Affiliations: }} \\ 
$^1$ Zhejiang University \\
\subsubsection*{\textit{HNU-VPAI} Team}
\noindent\textit{\textbf{Members: }} \\
Zhiyu Wang$^1$ (\mhref{mailto:zhiyuwang@hnu.edu.cn}{zhiyuwang@hnu.edu.cn})\\
Xudong Kang$^1$ (\mhref{mailto:xudong\_kang@hnu.edu.cn}{xudong\_kang@hnu.edu.cn}) \\
Shutao Li$^1$ (\mhref{mailto:shutao\_li@hnu.edu.cn}{shutao\_li@hnu.edu.cn})\\
\noindent\textit{\textbf{Affiliations: }} \\ 
$^1$ Hunan University \\
\subsubsection*{\textit{lababa} Team}
\noindent\textit{\textbf{Members: }} \\
Yumei Li$^1$ (\mhref{mailto:liym0302@163.com }{liym0302@163.com}) \\
Cong Xu$^1$ (\mhref{mailto:xucong143336@gmail.com}{xucong143336@gmail.com}) \\ 
Pu Luo$^1$ (\mhref{mailto:18581668812@163.com}{18581668812@163.com})\\
\noindent\textit{\textbf{Affiliations: }} \\
$^1$ Xidian University \\
\subsubsection*{\textit{Location depends on guessing} Team}
\noindent\textit{\textbf{Members: }} \\
Xin Wei$^1$ (\mhref{mailto:weix11@chinatelecom.cn}{weix11@chinatelecom.cn})\\
Han Fang$^1$ (\mhref{mailto:fangh2@chinatelecom.cn}{fangh2@chinatelecom.cn}) \\
Xiaodong Dong$^1$ (\mhref{mailto:dongxd1@chinatelecom.cn}{dongxd1@chinatelecom.cn}) \\
Hongbo Sun$^1$ (\mhref{mailto:sunhb3@chinatelecom.cn}{sunhb3@chinatelecom.cn}) \\
Mengxi Jia$^1$ (\mhref{mailto:jiamx1@chinatelecom.cn}{jiamx1@chinatelecom.cn}) \\
Ye Yuan$^1$ (\mhref{mailto:yuany2@chinatelecom.cn}{yuany2@chinatelecom.cn}) \\
Zhiyong Feng$^2$ (\mhref{mailto:24s136095@stu.hit.edu.cn}{24s136095@stu.hit.edu.cn}) \\
Tianyi Gao$^3$ (\mhref{mailto:tyigao@stu.xjtu.edu.cn}{tyigao@stu.xjtu.edu.cn}) \\
Yongkang Yu$^4$ (\mhref{mailto:yyk1249501542@gmail.com}{yyk1249501542@gmail.com}) \\
Haobo Cheng$^5$ (\mhref{mailto:chbustc@mail.ustc.edu.cn}{chbustc@mail.ustc.edu.cn}) \\
Muyang Yan$^6$ (\mhref{mailto:mario112112@163.com}{mario112112@163.com})\\
\noindent\textit{\textbf{Affiliations: }} \\ 
$^1$ Institute of Artificial Intelligence, China Telecom \\
$^2$ Harbin Institute of Technology \\
$^3$ Xi'an Jiaotong University \\
$^4$ Beijing University of Posts and Telecommunications \\
$^5$ University of Science and Technology of China \\
$^6$ East China Normal University \\
\subsubsection*{\textit{MILVLG\_HDU} Team}
\noindent\textit{\textbf{Members: }} \\
Zhenwei Shao$^1$ (\mhref{mailto:shaozw@hdu.edu.cn}{shaozw@hdu.edu.cn})\\
Lihao Zheng$^1$ (\mhref{mailto:zhenglh@hdu.edu.cn}{zhenglh@hdu.edu.cn})\\
Shuai Shao$^1$ (\mhref{mailto:242050206@hdu.edu.cn}{242050206@hdu.edu.cn}) \\
Zhou Yu$^1$ (\mhref{mailto:yuz@hdu.edu.cn}{yuz@hdu.edu.cn})\\
\noindent\textit{\textbf{Affiliations: }} \\ 
$^1$ Hangzhou Dianzi University, China\\
\subsubsection*{\textit{mm618} Team}
\noindent\textit{\textbf{Members: }} \\
Yanan Wang$^1$ (\mhref{mailto:wangyanan@mail.dlut.edu.cn}{wangyanan@mail.dlut.edu.cn})\\
Yicen Tian$^1$ (\mhref{mailto:yicentian@mail.dlut.edu.cn}{yicentian@mail.dlut.edu.cn}) \\
Dailin Li$^1$ (\mhref{mailto:ldlbest@mail.dlut.edu.cn}{ldlbest@mail.dlut.edu.cn}) \\
\noindent\textit{\textbf{Affiliations: }} \\ 
$^1$ Dalian University of Technology \\
\subsubsection*{\textit{NJUST--KMG} Team}
\noindent\textit{\textbf{Members: }} \\
Yipeng Lin$^1$ (\mhref{mailto:lllinyp@163.com}{lllinyp@163.com})\\
Yang Yang$^1$ (\mhref{mailto:yyang@njust.edu.cn}{yyang@njust.edu.cn})\\
\noindent\textit{\textbf{Affiliations: }} \\ 
$^1$ Nanjing University of Science and Technology \\
\subsubsection*{\textit{rookiesllm} Team}
\noindent\textit{\textbf{Members: }} \\
Huayong Hu$^1$ (\mhref{mailto:huhy6519@mails.jlu.edu.cn}{huhy6519@mails.jlu.edu.cn})\\
Yuxin Qin$^1$ (\mhref{mailto:qinyx23@mails.jlu.edu.cn}{qinyx23@mails.jlu.edu.cn}) \\
Shijie Dong$^2$ (\mhref{mailto:dsj20020312@163.com}{dsj20020312@163.com}) \\
Qi Li$^1$ (\mhref{mailto:qili23@mails.jlu.edu.cn}{qili23@mails.jlu.edu.cn})\\
\noindent\textit{\textbf{Affiliations: }} \\ 
$^1$ Jilin University \\
$^2$ Beijing Institute of Technology \\
\subsubsection*{\textit{SRCN-AIVL} Team}
\noindent\textit{\textbf{Members: }} \\
Cheng Chen$^1$ (\mhref{mailto:koalala.chen@samsung.com}{koalala.chen@samsung.com}) \\
Ziyang Peng$^1$ (\mhref{mailto:ziyang.peng@samsung.com}{ziyang.peng@samsung.com}) \\
Yin Tang$^1$ (\mhref{mailto:yin6913.tang@samsung.com}{yin6913.tang@samsung.com})\\
Jiang Yu$^1$ (\mhref{mailto:jiang0922.yu@samsung.com}{jiang0922.yu@samsung.com})\\
Weiyang Su$^1$ (\mhref{mailto:weiyang.su@samsung.com}{weiyang.su@samsung.com})\\
\noindent\textit{\textbf{Affiliations: }} \\ 
$^1$ Samsung Electronics (China) R\&D Center \\
\subsubsection*{\textit{Star\_s} Team}
\noindent\textit{\textbf{Members: }} \\
Yanglin Deng$^1$ (\mhref{mailto:yanglin\_deng@163.com}{yanglin\_deng@163.com}) \\
Jinglin Zhou$^1$ (\mhref{mailto:6233114044@stu.jiangnan.edu.cn}{6233114044@stu.jiangnan.edu.cn}) \\
Hongyao Chen$^1$ (\mhref{mailto:6233111020@stu.jiangnan.edu.cn}{6233111020@stu.jiangnan.edu.cn}) \\
Wei Zhang$^1$ (\mhref{mailto:phenixnull@gmail.com}{phenixnull@gmail.com}) \\
Xujie Zhou$^1$ (\mhref{mailto:zxj165561@gmail.com}{zxj165561@gmail.com}) \\
Tianyang Xu$^1$ (\mhref{mailto:tianyang.xu@jiangnan.edu.cn}{tianyang.xu@jiangnan.edu.cn}) \\
Xiao-Jun Wu$^1$ (\mhref{mailto:wu\_xiaojun@jiangnan.edu.cn}{wu\_xiaojun@jiangnan.edu.cn}) \\
Josef Kittler$^2$ (\mhref{mailto:j.kittler@surrey.ac.uk}{j.kittler@surrey.ac.uk}) \\
\noindent\textit{\textbf{Affiliations: }} \\ 
$^1$ Jiangnan University\\
$^2$ University of Surrey, UK\\
\subsubsection*{\textit{SUP} Team}
\noindent\textit{\textbf{Members: }} \\
Yashu Kang$^{1,2}$ (\mhref{mailto:kangyashu@supconit.com}{kangyashu@supconit.com})\\
Zhehao Shen$^{1,3}$ (\mhref{mailto:20234246028@stu.suda.edu.cn}{20234246028@stu.suda.edu.cn}) \\
Yuzhe Cen$^{1,4}$ (\mhref{mailto:yc4494@columbia.edu}{yc4494@columbia.edu})\\
\noindent\textit{\textbf{Affiliations: }} \\ 
$^1$ Zhejiang Supcon Information Technology \\
$^2$ Zhejiang University of Technology \\
$^3$ Soochow University \\
$^4$ Columbia University \\
\subsubsection*{\textit{Tang\_TUTE} Team}
\noindent\textbf{Members:} \\
Guopeng Tang$^1$ (\mhref{mailto:pastwill@163.com}{pastwill@163.com}) \\
\noindent\textbf{Affiliations:} \\
$^1$ \dcheck{Tianjin University of Technology and Education} \\
\subsubsection*{\textit{Tele\_AI} Team}
\noindent\textit{\textbf{Members: }} \\
Hui Li$^1$ (\mhref{mailto:lih@chinatelecom.cn}{lih@chinatelecom.cn})\\
Zhaofan Zou$^1$ (\mhref{mailto:zouzhf41@chinatelecom.cn}{zouzhf41@chinatelecom.cn})\\
Feiyu Wang$^2$ (\mhref{mailto:wfy\_0502@163.com}{wfy\_0502@163.com})\\
Yanbin Huang$^3$ (\mhref{mailto:thinkfulcat@gmail.com}{thinkfulcat@gmail.com}) \\
Yilin Tao$^4$ (\mhref{mailto:tyilin034725@163.com}{tyilin034725@163.com})\\
\noindent\textit{\textbf{Affiliations: }} \\ 
$^1$ Institute of Artificial Intelligence, China Telecom  \\
$^2$ Beijing Language and Culture University  \\
$^3$ Huazhong University of Science and Technology  \\
$^4$ University of Chinese Academy of Sciences  \\
\subsubsection*{\textit{WDL} Team}
\noindent\textit{\textbf{Members: }} \\
Zhenglin Du$^1$ (\mhref{mailto:zhenglin\_du@163.com}{zhenglin\_du@163.com})\\
Yi Wen$^1$ (\mhref{mailto:18261172307@163.com}{18261172307@163.com})\\
Zhengyang Li$^1$ (\mhref{mailto:li\_zhengyang1222@163.com}{li\_zhengyang1222@163.com})\\
\noindent\textit{\textbf{Affiliations: }} \\ 
$^1$ Xidian University \\
\subsubsection*{\textit{yqhhh} Team}
\noindent\textit{\textbf{Members: }} \\
Yiqing Wang$^1$ (\mhref{mailto:24171213882@stu.xidian.edu.cn}{24171213882@stu.xidian.edu.cn})\\
Jing He$^1$ (\mhref{mailto:24171213874@stu.xidian.edu.cn}{24171213874@stu.xidian.edu.cn}) \\
\noindent\textit{\textbf{Affiliations: }} \\ 
$^1$ Xidian University \\
\subsubsection*{\textit{ZLC} Team}
\noindent\textit{\textbf{Members: }} \\
Zihan Zhai$^1$ (\mhref{mailto:25171213969@stu.xidian.edu.cn}{25171213969@stu.xidian.edu.cn})\\
Tingting Li$^1$ (\mhref{mailto:25241215337@stu.xidian.edu.cn}{25241215337@stu.xidian.edu.cn}) \\
Yuying Chen$^1$ (\mhref{mailto:25171214015@stu.xidian.edu.cn}{25171214015@stu.xidian.edu.cn})\\
\noindent\textit{\textbf{Affiliations: }} \\
$^1$ Xidian University \\
\subsubsection*{\textit{zls123} Team}
\noindent\textit{\textbf{Members: }} \\
Xiaopeng Zhou$^1$ (\mhref{mailto:TeamLeaderEmail@xxx.xxx}{z2362910193@gmail.com})\\
Chaoyang Liao$^1$ (\mhref{mailto:cylio529@gmail.com}{cylio529@gmail.com}) \\
Zhilong Song$^1$ (\mhref{mailto:908194340@qq.com}{908194340@qq.com})\\
\noindent\textit{\textbf{Affiliations: }} \\ 
$^1$ Xidian University \\

\end{document}